\documentclass{article} 
\usepackage{iclr2020_conference,times}

\usepackage{amsmath,amsfonts,bm}

\def\eqref#1{equation~\ref{#1}}

\def\1{\bm{1}}

\def\mP{{\bm{P}}}

\DeclareMathAlphabet{\mathsfit}{\encodingdefault}{\sfdefault}{m}{sl}
\SetMathAlphabet{\mathsfit}{bold}{\encodingdefault}{\sfdefault}{bx}{n}

\usepackage[utf8]{inputenc} 
\usepackage[T1]{fontenc}    
\usepackage{hyperref}       
\usepackage{url}            
\usepackage{booktabs}       
\usepackage{amsfonts}       
\usepackage{nicefrac}       
\usepackage{microtype}      

\usepackage{subcaption}
\usepackage{multirow}

\title{Multi-Scale Representation Learning  for Spatial Feature Distributions using Grid Cells}

\author{
Gengchen Mai$^1$, Krzysztof Janowicz$^1$, Bo Yan$^2$, Rui Zhu$^1$, Ling Cai$^1$ \& Ni Lao$^3$ \\
$^1$STKO Lab, University of California, Santa Barbara, CA, USA, 93106 \\
\texttt{\{gengchen\_mai,janowicz,ruizhu,lingcai\}@ucsb.edu} \\
$^2$LinkedIn Corporation, Mountain View, CA, USA, 94043 \\
\texttt{boyan1@linkedin.com} \\
$^3$SayMosaic Inc., Palo Alto, CA, USA, 94303 \\
\texttt{ni.lao@mosaix.ai} 
}

\usepackage{amsthm}
\usepackage{amsmath}
\usepackage{mathabx}
\usepackage{xspace}
\usepackage{color}
\usepackage{pgfplots, pgfplotstable}
\usepackage[colorinlistoftodos]{todonotes} 

\newcommand{\comment}[1]{}

\renewcommand{\vec}[1]{\boldsymbol{\mathbf{#1}}}

\def\bm{\vec{m}}

\def\bv{\vec{v}}

\def\bx{\vec{x}}
\def\by{\vec{y}}

\def\mP{\mathcal{P}}
\def\Real{\mathbb{R}}
\def\Complex{\mathbb{C}}

\def\int{\mathrm{int}}

\newtheorem{theorem}{Theorem}

\newcommand{\modelname}{Space2Vec }
\newcommand{\model}[2]{{#1}}

\newcommand{\leakyrelu}{LeakyReLU}

\newcommand{\loss}{\mathcal{L}}

\newcommand{\emb}{\mathbf{e}}

\newcommand{\neisize}{n}

\newcommand{\spdim}{L}
\newcommand{\nscale}{S}

\newcommand{\pedim}{d^{(x)}}
\newcommand{\fedim}{d^{(v)}}
\newcommand{\embdim}{d}

\newcommand{\peemb}[1]{\mathbf{e}[\bx_{#1}]}
\newcommand{\feemb}[1]{\mathbf{e}[\bv_{#1}]}

\newcommand{\negsize}{N}
\newcommand{\negsamp}{\mathcal{N}}
\newcommand{\negsampi}{\negsamp_i}

\newcommand{\attnscr}{\alpha}
\newcommand{\attnvec}{\mathbf{a}}
\newcommand{\attnsize}{K}
\newcommand{\attnidx}{k}
\newcommand{\attnact}{g} 

\newcommand{\freq}{S}

\newcommand{\enc}{Enc}
\newcommand{\enctheta}{\mathbf{\theta}_{\text{enc}}}

\newcommand{\vsection}[1]{
\vspace{-0.2cm}
\section{#1}
\vspace{-0.2cm}
}
\newcommand{\vsubsection}[1]{
\vspace{-0.2cm}
\subsection{#1}
\vspace{-0.2cm}
}

\newcommand{\vsubsubsection}[1]{
\vspace{-0.05cm}
\subsubsection{#1}
\vspace{-0.1cm}
}

\newcommand{\dectheta}{\mathbf{\theta}_{\text{dec}}}

\newcommand{\locdec}{Dec_s}
\newcommand{\locdectheta}{\mathbf{\theta}_{\text{dec}_s}}

\newcommand{\contdec}{Dec_c}
\newcommand{\contdectheta}{\mathbf{\theta}_{\text{dec}_c}}

\newcommand{\unit}{\mathbf{a}}
\newcommand{\spfuc}{\phi}
\newcommand{\theofuc}{\mathbf{\Psi}}
\newcommand{\theomat}{\mathbf{C}}

\newcommand{\pemlp}{\mathbf{NN}}

\newcommand{\locdecmlp}{\mathbf{NN}_{\text{dec}}}

\newcommand{\loc}{loc}
\newcommand{\cont}{cont}

\newcommand{\aodha}{wrap}
\newcommand{\naive}{direct}

\newcommand{\numresnet}{h}
\newcommand{\numneuron}{o}

\iclrfinalcopy 
\begin{document}

\maketitle

\begin{abstract}
  Unsupervised text encoding models 
  have recently fueled substantial progress in Natural Language Processing (NLP).
  The key idea is to use neural networks to convert words in texts to vector space representations (embeddings) based on word positions in a sentence and their contexts, which are suitable for end-to-end training of downstream tasks. We see a strikingly similar situation in 
  spatial analysis, which focuses on incorporating both absolute positions and spatial contexts of geographic objects such as Points of Interest (POIs) into models. 
  A general-purpose representation model for space   is valuable for a multitude of  tasks.
  However, no such general model exists to date beyond 
  simply applying discretization or feed-forward nets to coordinates,
  and little effort has been put into jointly modeling distributions with vastly different characteristics,  which commonly emerges from GIS data.
  %
  Meanwhile, Nobel Prize-winning Neuroscience research shows that grid cells in mammals provide a multi-scale periodic representation that functions as a metric for location encoding and is critical for recognizing places and for path-integration. 
  Therefore, we propose a representation learning model called \modelname to encode the absolute positions and spatial relationships of places. %
We conduct experiments on two real-world geographic data for two different tasks: 1) predicting types of POIs given their positions and context, 2) image classification
leveraging their geo-locations.
Results show that  because of its multi-scale representations, \modelname 
outperforms well-established ML approaches such as RBF kernels, multi-layer feed-forward nets, and tile embedding approaches for location modeling and image classification tasks. 
Detailed analysis  shows that all baselines can at most well handle distribution at one scale but show poor performances in other scales. In contrast, \modelname’s multi-scale representation can handle distributions at different scales.
  %
  \footnote{Link to project repository: \url{https://github.com/gengchenmai/space2vec}}
\end{abstract}

\vsection{Introduction}     \label{sec:intro}

Unsupervised text encoding models such as Word2Vec~\citep{mikolov2013distributed}, Glove~\citep{pennington2014glove}, ELMo~\citep{peters2018deep}, and BERT~\citep{devlin2018bert} have been effectively utilized in many Natural Language Processing (NLP) tasks. 
At their core they
train models which encode words into vector space representations based on their positions in the text and their context. 
A similar situation can be encountered in the field of Geographic Information Science (GIScience). For example, spatial interpolation aims at predicting an attribute value, e.g., elevation,  at an unsampled location based on the known attribute values of nearby samples. 
Geographic information has become an important component to many  tasks such as fine-grained image classification~\citep{mac2019presence}, 
point cloud classification and semantic segmentation~\citep{qi2017pointnet}, 
reasoning about Point of Interest (POI) type similarity ~\citep{yan2017itdl}, 
land cover classification \citep{kussul2017deep},
and geographic question answering \citep{mai2019relaxing}.
Developing a \textit{general} model for 
vector space representation of any point in space
would pave the way for many future applications.

\begin{figure*}[h!]
	\centering
	\begin{subfigure}[b]{0.19\textwidth}
		\centering
		\includegraphics[width=\textwidth]{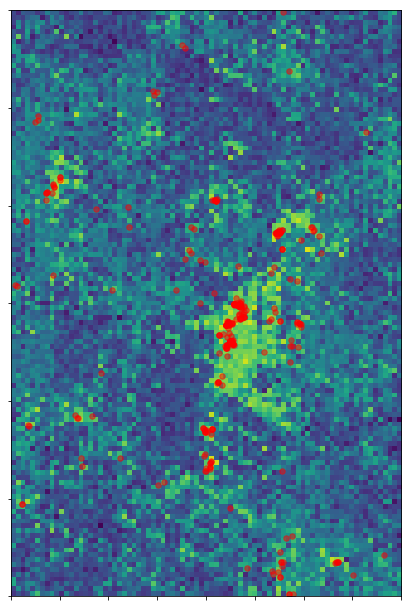}
		\subcaption[]{Women's Cloth}
		\label{fig:women_cloth}
	\end{subfigure}
	\begin{subfigure}[b]{0.19\textwidth}
		\centering
		\includegraphics[width=\textwidth]{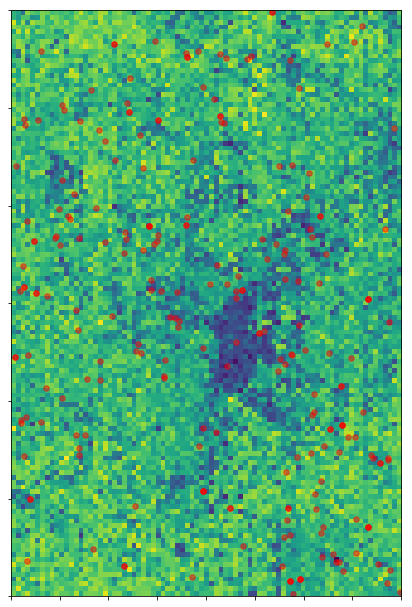}
		\subcaption[]{Education
		}
		\label{fig:education}
	\end{subfigure}
	\begin{subfigure}[b]{0.282\textwidth}  
		\centering 
		\includegraphics[width=\textwidth]{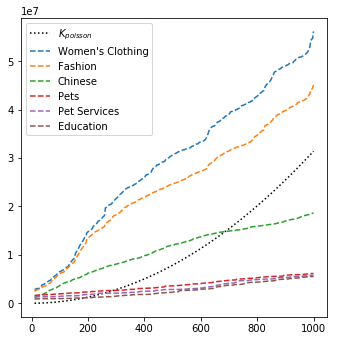}
		\subcaption[]{Ripley's K
		}
		\label{fig:ripley}
	\end{subfigure}
	\vspace*{-0.05cm}
	\begin{subfigure}[b]{0.31\textwidth}
		\centering
		\includegraphics[width=\textwidth]{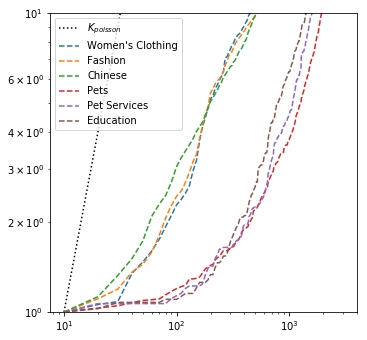} 
		\subcaption[]{Renormalized Ripley's K
		}
		\label{fig:ripley_log}
	\end{subfigure}

	\vspace*{-0.05cm}
	\caption{\small
	The challenge of joint modeling distributions with very different characteristics.
	(a)(b)  The POI locations (red dots) in Las Vegas and \modelname predicted conditional likelihood of 
	Women's Clothing (with a clustered distribution) and 
    Education (with an even distribution). The dark area in (b) indicates that the downtown area has more POIs of  other types   than  education.
	(c)  Ripley's K curves of POI types for which \modelname has the largest and smallest improvement over  $\aodha$ \citep{mac2019presence}. Each curve  represents  the number of POIs of a certain type inside certain radios centered at every POI of that type; 
	(d)  Ripley's K curves renormalized by POI densities and shown in log-scale. 
	To efficiently achieve multi-scale representation \modelname concatenates the grid cell encoding of 64 scales (with wave lengths ranging from 50 meters to $40k$ meters) as the first layer of a deep model, and trains with POI data in an unsupervised fashion.
	}
	\label{fig:motivation}
	\vspace*{-0.5cm}
\end{figure*}

However, existing models often utilize \textit{specific} methods to deal with geographic information  and often disregards geographic coordinates. 
For example, Place2Vec~\citep{yan2017itdl} converts the coordinates of POIs into spatially collocated POI pairs within certain distance bins, and does not preserve information about the (cardinal) direction between points. 
\citet{li2017diffusion} propose 
DCRNN for traffic forecasting in which the traffic sensor network is converted to a distance weighted graph which necessarily forfeits information about the spatial layout of sensors.
There is, however, no general representation  model 
 beyond simply applying discretization  \citep{berg2014birdsnap, tang2015improving}  or feed-forward nets \citep{chu2019geo,mac2019presence} to coordinates.


A key challenge in developing a general-purpose representation model for space is how to deal with mixtures of distributions with very different characteristics (see an example in  Figure~\ref{fig:motivation}), which often emerges in spatial datasets~\citep{mckenzie2015poi}.
For example, there are POI types with clustered distributions such as women's clothing, 
while there are other POI types with regular distributions such as education. 
These feature distributions co-exist in the same space,  and yet we want a single  representation to accommodate all of them in a task such as location-aware image classification~\citep{mac2019presence}.
%
Ripley's K is a spatial analysis method used to describe point patterns over a given area of interest.
Figure \ref{fig:ripley} shows the K plot of several POI types in Las Vegas. One can see that as the radius grows the numbers of POIs increase at different rates for different  POI types.
In order to see the relative change of density at different scales, 
we renormalize the curves by each POI type's density and show it in log scale in Figure~\ref{fig:ripley_log}.
One can see two distinct POI type groups with different distribution patterns with
clustered and even distributions. 
If we want to model the distribution
of these POIs by discretizing the study area into tiles, 
we have to use small grid sizes for women's clothing
while using larger grid sizes for educations because smaller grid sizes lead to over-	parameterization of the model and overfitting. 
In order to 
jointly describe these distributions and their patterns, we need an encoding method which supports \textit{multi-scale representations}.

Nobel Prize winning Neuroscience research~\citep{abbott2014nobel} has demonstrated that grid cells in mammals provide a multi-scale periodic representation that functions as a metric for location encoding, which is critical for integrating self-motion. 
Moreover, \citet{blair2007scale} show that the multi-scale periodic representation of grid cells can be simulated by summing three cosine grating functions oriented $60^{\circ}$ apart, which may be regarded as a simple Fourier model of the hexagonal lattice. 
This research inspired us to encode locations with multi-scale periodic representations. 
Our assumption is that decomposed geographic coordinates 
helps 
machine learning models, such as deep neural nets,
and  multi-scale representations deal with the inefficiency of intrinsically single-scale methods such as RFB kernels or discretization (tile embeddings). 
To validate this intuition, we propose an encoder-decoder framework to encode the distribution of point-features\footnote{In GIS and spatial analysis, `features' are representations of real-world entities. A tree can, for instance, be modeled by a point-feature, while a street would be represented as a line string feature.} in space and train such a model in an unsupervised manner. This idea of using sinusoid functions with different frequencies to encode positions is similar to the position encoding proposed in the Transformer model~\citep{vaswani2017attention}. However, the position encoding model of Transformer deals with a discrete 1D space -- the positions of words in a sentence -- while our model works on higher dimensional continuous spaces such as the surface of earth. 

\textbf{In summary, the contributions of our work are as follows:}
\begin{enumerate}
	\item We propose an encoder-decoder encoding framework called \modelname using sinusoid functions with different frequencies to model 
	absolute positions and spatial contexts. 
	We also propose a multi-head attention mechanism
	based on context points. 
	To the best of our knowledge, this is the first attention model that explicitly considers the spatial relationships between the query point and context points.
	
	\item 
	We conduct experiments on two real world geographic data for two different tasks:
	1) predicting types of  POIs given their positions and context,
	2) image classification leveraging their geo-locations.
	\modelname 
	outperforms  well-established  encoding methods such as RBF kernels, multi-layer feed-forward nets, and tile embedding approaches for location modeling and image classification.
	
	\item To understand the advantages of \modelname
	we visualize the firing patterns (response maps) of location models' encoding layer neurons and show how they 
	handle spatial structures at different scales by integrating multi-scale representations.
	Furthermore the firing patterns for the 
	spatial context
	models neurons give insight into how the grid-like cells  
	capture the decreasing distance effect with multi-scale representations. 
\end{enumerate}

\vsection{Problem Formulation}  \label{sec:prob}

{\em Distributed representation of point-features in space} can be formulated as follows. 
Given a set of points $\mP=\{p_i\}$, i.e., Points of Interests (POIs), in $\spdim$-D space ($\spdim = 2,3$) 
define a function $f_{\mP,\theta}(\bx): \Real^{\spdim} \to \Real^\embdim$ ($\spdim \ll \embdim$), 
which is parameterized by $\theta$ and maps any coordinate $\bx$ in space to a vector representation of $\embdim$ dimension.
Each point (e.g., a restaurant) $p_i=(\bx_i, \bv_i)$ is associated with a location $\bx_i$ and  attributes $\bv_i$ (i.e., POI features such as type, name, capacity, etc.). 
The function $f_{\mP,\theta}(\bx)$ encodes the probability distribution of point features over space and can give a representation of any point in the space. Attributes (e.g. place types such as \textit{Museum}) and coordinate of point can be seen as analogies to words and word positions in 
commonly used
word embedding models. 

\vspace{-0.2cm}
\section{Related Work}   \label{sec:related}
\vspace{-0.2cm}

There has been theoretical research on neural network based path integration/spatial localization models and their relationships with grid cells. 
Both \citet{cueva2018emergence} and \citet{banino2018vector} showed that grid-like spatial response patterns emerge in trained networks for navigation tasks which demonstrate that grid cells are critical for vector-based navigation. 
Moreover, \citet{gao2018learning} propose a representational model for grid cells in navigation tasks which has good quality such as magnified local isometry. All these research is focusing on understanding the relationship between the grid-like spatial response patterns and navigation tasks from a theoretical perspective. In contrast, our goal focuses on utilizing these theoretical results
on real world data in geoinformatics.

Radial Basis Function (RBF) kernel is a well-established approach to generating learning friendly representation  
from points in space  
for machine learning 
algorithms such as SVM classification~\citep{baudat2011kernel} and regression~\citep{Bierens1994regression}.
However, the representation is example based
-- i.e., the resultant model uses the positions of training examples as the centers of Gaussian kernel functions~\citep{mazya1996gaussian}.
In comparison, the grid cell based location encoding relies on sine and cosine functions, and  the resultant model is inductive and does not 
store training examples.

Recently the computer vision community shows increasing interests in incorporating geographic information (e.g. coordinate encoding) into neural network architectures for multiple tasks such as image classification \citep{tang2015improving} and fine grained recognition \citep{berg2014birdsnap,chu2019geo,mac2019presence}. 
Both \citet{berg2014birdsnap} and \citet{tang2015improving} proposed to discretize the study area into regular grids. To model the geographical prior distribution of the image categories, the grid id is used for GPS encoding instead of the raw coordinates. However, choosing the correct discretization is challenging \citep{openshaw1984modifiable,fotheringham1991modifiable}, and incorrect choices can significantly affect the final performance \citep{moat2018refining,lechner2012investigating}. 
In addition, discretization does not scale well in terms of memory use. To overcome these difficulties, both \citet{chu2019geo} and \citet{mac2019presence} advocated the idea of inductive location encoders which directly encode coordinates into a location embedding. However, both of them directly feed the coordinates into a feed-forward neural network \citep{chu2019geo} or residual blocks \citep{mac2019presence} without any feature decomposition strategy. 
Our experiments show that this direct  encoding approach is insufficient to capture the spatial feature distribution and \modelname significantly outperforms them by integrating spatial representations of different scales.

\vsection{Method} \label{sec:method}

We solve \textit{distributed representation of point-features in space} 
(defined in Section \ref{sec:prob}) with  an encoder-decoder architecture:
\begin{enumerate}

\item Given a point $p_i=(\bx_i, \bv_i)$ a \textbf{point space encoder} $\enc^{(x)}()$ encodes location $\bx_i$ into a location embedding $\peemb{i} \in \Real^{\pedim}$ and a \textbf{point feature encoder} $\enc^{(v)}()$ encodes its feature into a feature embedding $\feemb{i} \in \Real^{\fedim}$.  
$\emb = [\peemb{i};\feemb{i}] \in \Real^{\embdim}$ is the full representation of point $p_i \in \mP$, where $\embdim = \pedim+\fedim$. 
$[;]$ represents vector concatenation. 
In contrast, geographic entities not in $\mP$ within the studied space can be represented by their location embedding $\peemb{j}$ since  
its  $\bv_i$ is unknown.
\item 
We developed two types of decoders which can be used independently or jointly.
A \textbf{location decoder} $\locdec()$  reconstructs point feature embedding $\feemb{i}$ given location embedding $\peemb{i}$, and 
a \textbf{spatial context decoder} $\contdec()$ reconstructs the feature embedding $\feemb{i}$ of point $p_i$ based on the space  and feature embeddings $\{\emb_{i1},...,\emb_{ij},...,\emb_{i\neisize}\}$ of 
nearest neighboring points $\{p_{i1},...,p_{ij},...,p_{i\neisize}\}$, where $\neisize$  is a hyper-parameter. 
\end{enumerate}

\vsubsection{Encoder}  
\label{sec:encoder}

\paragraph{Point Feature Encoder}          \label{sec:feenc}
Each point $p_i=(\bx_i, \bv_i)$ in a point set $\mP$ is often associated with  features  such as the air pollution station data associate with some air quality measures, a set of POIs with POI types and names, a set of points from survey and mapping with elevation values, a set of points from geological survey with mineral content measure, and so on.  
The point feature encoder $\enc^{(v)}()$ encodes such features $\bv_i$ 
into a feature embedding $\feemb{i} \in \Real^{\fedim}$. The implementation of $\enc^{(v)}()$ depends on the nature of these features. For example, if each point represents a POI with multiple POI types (as in this study), the feature embedding $\feemb{i}$ can simply be the mean of each POI types' embeddings 
$
\feemb{i} = \dfrac{1}{H} \sum_{h=1}^{H} \mathbf{t}_{h}^{(\gamma)},
$
where $\mathbf{t}_{h}^{(\gamma)}$ indicates the $h$th POI type embedding of a POI $p_i$ with $H$ POI types.
We apply $L_{2}$ normalization 
to the POI type embedding matrix.

\paragraph{Point Space Encoder}  \label{sec:peenc}
A part of the novelty of this paper is from the point space encoder $\enc^{(x)}()$. 
We first introduce Theorem \ref{the:gc} which provide an analytical solution $\spfuc(\bx)$ as the base of encoding any location $\bx \in \Real^2$ in 2D space to a distributed representation: 
\begin{theorem}
	\label{the:gc}
	Let $\theofuc(\bx) = (e^{i\langle \unit_j, \bx \rangle}, j=1,2,3)^T \in \Complex^3$ where $e^{i\theta} = \cos\theta + i\sin\theta$ is the Euler notation of complex values; $\langle \unit_j, \bx \rangle$ is the inner product of $\unit_j$ and $\bx$. $\unit_1, \unit_2, \unit_3 \in \Real^2$ are 2D vectors such that the angle between $\unit_k$ and $\unit_l$ is $2\pi/3$, $\forall j, \|\unit_j\| = 2\sqrt{\alpha}$. Let $\theomat \in \Complex^{3 \times 3}$ be a random complex matrix such as $\theomat^{*}\theomat = \mathbf{I}$. Then $\spfuc(\bx) = \theomat\theofuc(\bx)$, $M(\Delta\bx) = \theomat diag(\theofuc(\Delta\bx))\theomat^{*}$ satisfies 
	\begin{equation}
		\spfuc(\bx+\Delta\bx) = M(\Delta\bx)\spfuc(\bx)
		\label{equ:theo1}\\
	\end{equation} and
	\begin{equation}
	\langle \spfuc(\bx+\Delta\bx), \spfuc(\bx) \rangle = d(1-\alpha\|\Delta\bx\|^2)
	\label{equ:theo2}
	\end{equation}
	where $d=3$ is the dimension of $\spfuc(\bx)$ and $\Delta\bx$ is a small displacement from $\bx$.
\end{theorem}
The proof of Theorem \ref{the:gc} can be seen in
\citet{gao2018learning}. $\spfuc(\bx) = \theomat\theofuc(\bx) \in \Complex^3$ amounts to a 6-dimension real value vector and each dimension shows a \textit{hexagon} firing pattern which models the grid cell behavior. 
Because of the periodicity of $sin()$ and $cos()$, this single scale representation  $\spfuc(\bx)$ does not form a global codebook of 2D positions, i.e. there can be $\bx \neq \by$, but $\spfuc(\bx) = \spfuc(\by)$. 

Inspired by Theorem \ref{the:gc} and the multi-scale periodic representation of grid cells in mammals~\citep{abbott2014nobel}
we set up our point space encoder $\peemb{} = \enc_{theory}^{(x)}(\bx)$ to use sine and cosine functions of different frequencies to encode positions in space.
Given any point $\bx$ in the studied 2D space, 
the space encoder $\enc_{theory}^{(x)}(\bx) = \pemlp(PE^{(t)}(\bx))$ 
where $PE^{(t)}(\bx) = [PE^{(t)}_{0}(\bx);...;PE^{(t)}_{s}(\bx);...;PE^{(t)}_{\nscale-1}(\bx)]$
is a concatenation of multi-scale representations of $\pedim= 6\nscale$ dimensions. 
Here $\nscale$ is the total number of grid scales and $s = 0,1,2,...,{\nscale-1}$.
$\pemlp()$ 
represents fully connected ReLU layers.
Let $\unit_1 = [1,0]^T, \unit_2 = [-1/2,\sqrt{3}/2]^T, \unit_3 = [-1/2,-\sqrt{3}/2]^T \in \Real^2$ be three unit vectors and the angle between any of them is $2\pi/3$. 
$\lambda_{min}, \lambda_{max}$ are the minimum and maximum grid scale and $g = \frac{\lambda_{max}}{\lambda_{min}}$.
At each  scale $s$, $PE^{(t)}_{s}(\bx) = [PE^{(t)}_{s,1}(\bx);PE^{(t)}_{s,2}(\bx);PE^{(t)}_{s,3}(\bx)]$ is a concatenation of three components, where
\begin{align}
PE^{(t)}_{s,j}(\bx) = [
\cos(\frac{\langle \bx, \unit_{j} \rangle}{ \lambda_{min} \cdot g^{s/(\nscale-1)}}); 
\sin(\frac{\langle \bx, \unit_{j} \rangle}{ \lambda_{min} \cdot g^{s/(\nscale-1)}})] \forall j = 1,2,3; 
\label{equ:petheory}
\end{align}

$\pemlp()$ and $PE^{(t)}(\bx)$ are analogies of $\theomat$ and $\theofuc(\bx)$ in Theorem \ref{the:gc}.

Similarly we can define another space encoder $\enc_{grid}^{(x)}(\bx) = \pemlp(PE^{(g)}(\bx))$  inspired by the position encoding model of Transformer \citep{vaswani2017attention}, where $PE^{(g)}(\bx) = [PE^{(g)}_{0}(\bx);...;PE^{(g)}_{s}(\bx);...;PE^{(g)}_{\nscale-1}(\bx)]$ is still a concatenation of its multi-scale representations, while $PE^{(g)}_{s}(\bx)=[PE^{(g)}_{s,1}(\bx);PE^{(g)}_{s,2}(\bx)]$ 
handles 
each component $l$ of $\bx$ 
separately:
\begin{align}
PE^{(g)}_{s,l}(\bx) = [
\cos(\frac{\bx^{[l]}}{ \lambda_{min} \cdot g^{s/(\nscale-1)} }); 
\sin(\frac{\bx^{[l]}}{ \lambda_{min} \cdot g^{s/(\nscale-1)} })] \forall l = 1,2
\label{equ:pegrid}
\end{align}


\vspace{-0.5cm}
\vsubsection{Decoder}   \label{sec:dec} 
Two types of decoders are designed for two major types of GIS problems: location modeling and spatial context modeling (See Section \ref{sec:dataset}).
\paragraph{Location Decoder}            \label{sec:locdec}
 $\locdec()$ directly reconstructs point feature embedding $\feemb{i}$ given its space embedding $\peemb{i}$. We use one layer feed-forward neural network $\locdecmlp()$ 
\begin{align}
\feemb{i}^{\prime} = \locdec( \bx_i; \locdectheta) = \locdecmlp(\peemb{i})
\label{equ:decloc}
\end{align}
For training we use inner product to compare the reconstructed feature embedding $\feemb{i}^{\prime}$ against the real feature embeddings of $\feemb{i}$ and other negative points (see training detail in Sec~\ref{sec:training}).

\paragraph{Spatial Context Decoder}       \label{sec:contdec}
$\contdec()$ reconstructs the feature embedding $\feemb{i}$ of the center point $p_i$ based on the space and feature embeddings $\{\emb_{i1},...,\emb_{ij},...,\emb_{i\neisize}\}$ of $\neisize$ nearby points $\{p_{i1},...,p_{ij},...,p_{i\neisize}\}$. 
Note that the feed-in order of context points 
 should not affect the prediction results, which can be achieved by 
permutation invariant neural network architectures \citep{zaheer2017deep} like PointNet \citep{qi2017pointnet}. 
\begin{align}
\feemb{i}^{\prime} = \contdec(\bx_i,
\{\emb_{i1},...,\emb_{ij},...,\emb_{i\neisize}\} ; \contdectheta) 
= \attnact(\dfrac{1}{\attnsize} \sum_{\attnidx=1}^{\attnsize} \sum_{j=1}^{\neisize} \attnscr_{ij\attnidx}\feemb{ij}) 
\label{equ:dec_gc}
\end{align}
Here $\attnact$ is an activation function such as sigmoid. $\attnscr_{ij\attnidx}=  \frac{exp(\sigma_{ijk})}{\sum_{o=1}^{\neisize} exp(\sigma_{iok})}$ is the attention of $p_{i}$
with its $j$th neighbor through the $k$th attention head, and 
\begin{align}
\sigma_{ijk} = \leakyrelu(\attnvec_k^{T}[\feemb{i}_{init}; \feemb{ij};\mathbf{e}[\bx_{i}-\bx_{ij}] ])
\label{equ:atten_gc}
\end{align}
where $\attnvec_{\attnidx} \in \Real^{2\fedim+\pedim}$ is the attention parameter in the $\attnidx$th attention head. 
The multi-head attention mechanism is inspired by Graph Attention Network \citep{velivckovic2017graph} and \citet{mai2019contextual}. 

To represent the spatial relationship (distance and direction) 
between each context point $p_{ij} = (\bx_{ij}, \bv_{ij})$ and the center point $p_i = (\bx_i, \bv_i)$,
we use the space encoder $\enc^{(x)}()$ to encode the displacement between them $\Delta\bx_{ij} = \bx_{i}-\bx_{ij}$. 
Note that we are 
modeling the spatial interactions between the center point and  $\neisize$ context points simultaneously. 

In Eq. \ref{equ:atten_gc}, $\feemb{i}_{init}$ indicates the initial guess of the feature embedding $\feemb{i}$ of point $p_i$ which is computed by using another multi-head attention layer as Eq. \ref{equ:dec_gc} where the weight $\attnscr_{ij\attnidx}^{\prime}=  \frac{exp(\sigma_{ijk}^{\prime})}{\sum_{o=1}^{\neisize} exp(\sigma_{iok}^{\prime})}$. Here, $\sigma_{ijk}^{\prime}$ is computed as Eq. \ref{equ:atten_init} where the query embedding $\feemb{i}$ is excluded.

\vspace{-0.4cm}
\begin{align}
	\sigma_{ijk}^{\prime} = \leakyrelu(\attnvec_{k}^{\prime T}[ \feemb{ij};\mathbf{e}[\bx_{i}-\bx_{ij}] ])
	\label{equ:atten_init}
\end{align}

\vspace{-0.4cm}
\vsubsection{Unsupervised Training}  \label{sec:training}
The unsupervised learning task can simply be
maximizing the log likelihood of observing the true point $p_i$ at position $\bx_i$  among all the points in $\mP$ 
\begin{align}
\label{equ:loss_gc}
\centering
\loss_\mP(\theta) = - 
\sum_{p_{i} \in \mP} \log  P(p_i|p_{i1},...,p_{ij},...,p_{i\neisize}) 
= - 
\sum_{p_{i} \in \mP} \log \dfrac{
	\exp(\feemb{i}^{T}\feemb{i}^{\prime})}{\sum_{p_{o} \in \mP} 
	\exp(\feemb{o}^{T}\feemb{i}^{\prime})}
\end{align}
Here only the feature embedding of $p_i$ is used (without location embedding) to prevent revealing the identities of the point candidates, and $\theta=[ \enctheta;  \dectheta]$

Negative sampling by \citet{mikolov2013distributed} can be used to improve the efficiency of training
\begin{align}
\label{equ:loss_neg_gc}
\centering
\loss_\mP'(\theta) &= - 
\sum_{p_{i} \in \mP} \Big( 
\log \sigma(\feemb{i}^{T}\feemb{i}^{\prime}) + 
\dfrac{1}{|\negsampi|}\sum_{p_{o} \in \negsampi} 
\log \sigma(-\feemb{o}^{T}\feemb{i}^{\prime})  
\Big)
\end{align}
Here $\negsampi \subseteq \mP$ is a set of 
sampled negative points for $p_{i}$  
($p_{i} \notin \negsampi$) and $\sigma(x) = 1/(1 + e^{-x})$.

\vsection{Experiment}   \label{sec:exp}
In this section we compare \modelname with commonly used position encoding methods, and analyze them both quantitatively and qualitatively. 
\paragraph{Baselines}
Our baselines include
1) $\model{\naive}{loc}$  directly applying feed-forward nets \citep{chu2019geo};
2) $\model{tile}{loc}$ discretization \citep{berg2014birdsnap,adams2015frankenplace,tang2015improving};
3) $\model{\aodha}{loc}$
feed-forward nets with coordinate wrapping
 \citep{mac2019presence}; and 
4) $\model{rbf}{loc}$ Radial Basis Function (RBF) kernels \citep{baudat2011kernel,Bierens1994regression}.
See Appendix~\ref{sec:baselines} for details of the baselines.

\vsubsection{POI Type Classification Tasks} \label{sec:poi_type_class}
\paragraph{Dataset and Tasks}\label{sec:dataset}
To test the proposed model, we conduct experiments on geographic datasets with POI position and type information. 
We utilize the open-source dataset published by Yelp Data Challenge and select all POIs within the Las Vegas downtown area\footnote{The geographic range is (35.989438, 36.270897) for latitude and (-115.047977, -115.3290609) for longitude.}. 
There are 21,830 POIs with 1,191 different POI types 
in this dataset. 
Note that each POI may be associated with one or more types, and we do not use any other meta-data such as business names, reviews for this study.
We  project  geographic coordinates into projection coordinates  using the NAD83/Conus Albers projection coordinate system\footnote{\url{https://epsg.io/5070-1252}}. 
The POIs are split into training, validation, and test dataset with ratios 80\%:10\%:10\%. 
We create two tasks setups which represent different types of modeling need in 
Geographic Information Science:
\begin{itemize}
\item \textbf{Location Modeling} 
predicts 
the feature information associated with a POI
based on its location $\bx_i$ represented by the \textit{location decoder} $\locdec()$. 
This represents a large number of location prediction problems such as image fine grained recognition with geographic prior \citep{chu2019geo}, and 
species potential distribution prediction \citep{zuo2008geosvm}.
\item  \textbf{Spatial Context Modeling} 
predicts 
the feature information associated with a POI
based on its context 
$\{\emb_{i1},...,\emb_{ij},...,\emb_{i\neisize}\}$ represented by the \textit{spatial context decoder} $\contdec()$. 
This represents a collections of spatial context prediction problem such as spatial context based facade image classification \citep{yan2018xnet+}, and all spatial interpolation problems.
\end{itemize}


We use POI  prediction metrics to evaluate these models. Given the real point feature embedding $\feemb{i}$ and $\negsize$ negative feature embeddings $\negsampi = \{\feemb{i}^-\}$, we compare the predicted $\feemb{i}^{\prime}$ with them by cosine distance. The cosine scores are used to rank $\feemb{i}$ and $\negsize$ negative samples. The negative feature embeddings are the feature embeddings of points $p_j$ randomly sampled from $\mP$ and $p_i \neq p_j$. 
We evaluate each model using Negative Log-Likelihood (NLL), Mean Reciprocal Rank (MRR) and HIT@5 (the chance of the true POI being ranked to top 5.
We train and test each model 10 times to estimate standard deviations. 
See Appendix~\ref{sec:parameters} for hyper-parameter selection details.

\vsubsubsection{Location Modeling Evaluation} \label{sec:loceval}

\begin{figure*}[h!]
	\centering \tiny
	\vspace*{-0.3cm}
	\begin{subfigure}[b]{0.14\textwidth}  
		\centering 
		\includegraphics[width=\textwidth]{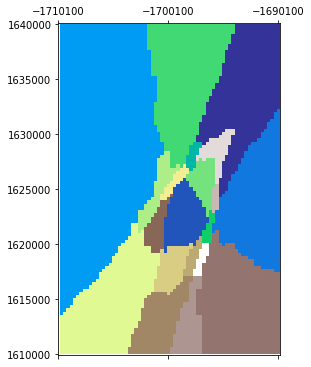}\vspace*{-0.1cm}
		\caption[]%
		{{\small 
		$\model{\naive}{loc}$
		}}    
		\label{fig:naivecluster}
	\end{subfigure}
	\hfill
	\begin{subfigure}[b]{0.137\textwidth}  
		\centering 
		\includegraphics[width=\textwidth]{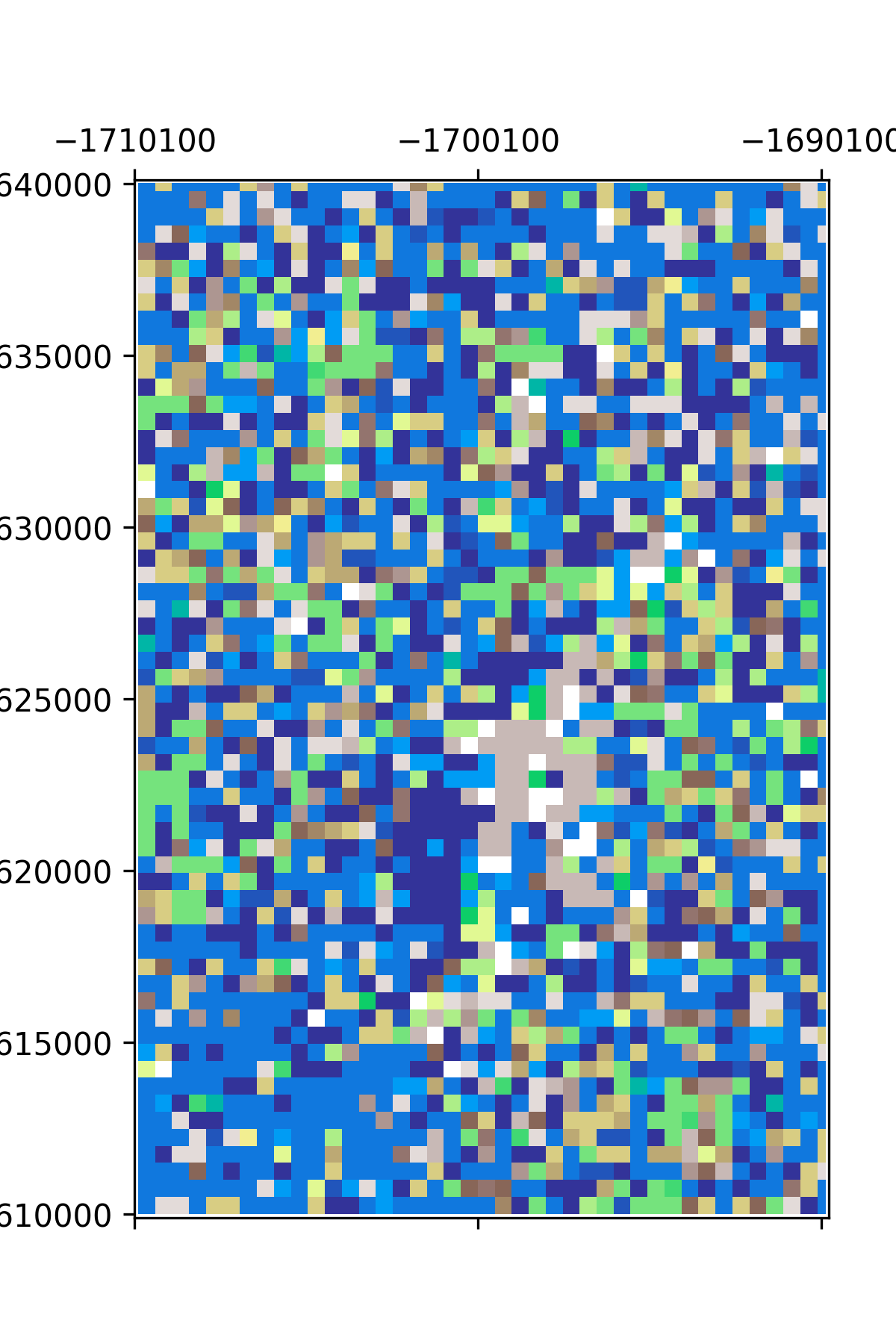}\vspace*{-0.4cm}
		\caption[]%
		{{\small 
		$\model{tile}{loc}$ 
		}}    
		\label{fig:tilecluster}
	\end{subfigure}
	\hfill
	\begin{subfigure}[b]{0.141\textwidth}  
		\centering 
		\includegraphics[width=\textwidth]{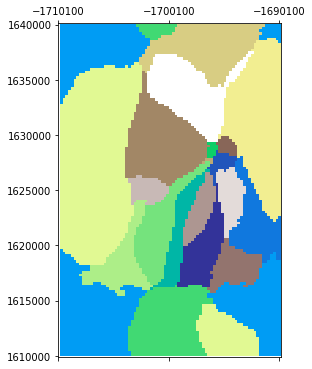}\vspace*{-0.1cm}
		\caption[]%
		{{\small 
		$\model{\aodha}{loc}$ 
		}}    
		\label{fig:aodhacluster}
	\end{subfigure}
	\hfill
	\begin{subfigure}[b]{0.145\textwidth}  
		\centering 
		\includegraphics[width=\textwidth]{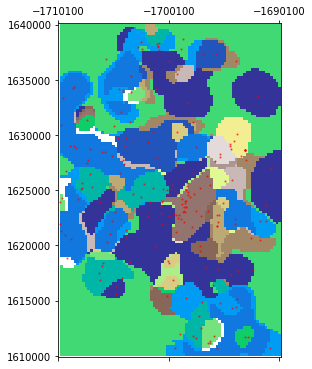}\vspace*{-0.15cm}
		\caption[]%
		{{\small 
		$\model{rbf}{loc}$ ($\sigma$=1k)
		}}    
		\label{fig:rbfcluster}
	\end{subfigure}
	\hfill
	\begin{subfigure}[b]{0.136\textwidth}  
		\centering 
		\includegraphics[width=\textwidth]{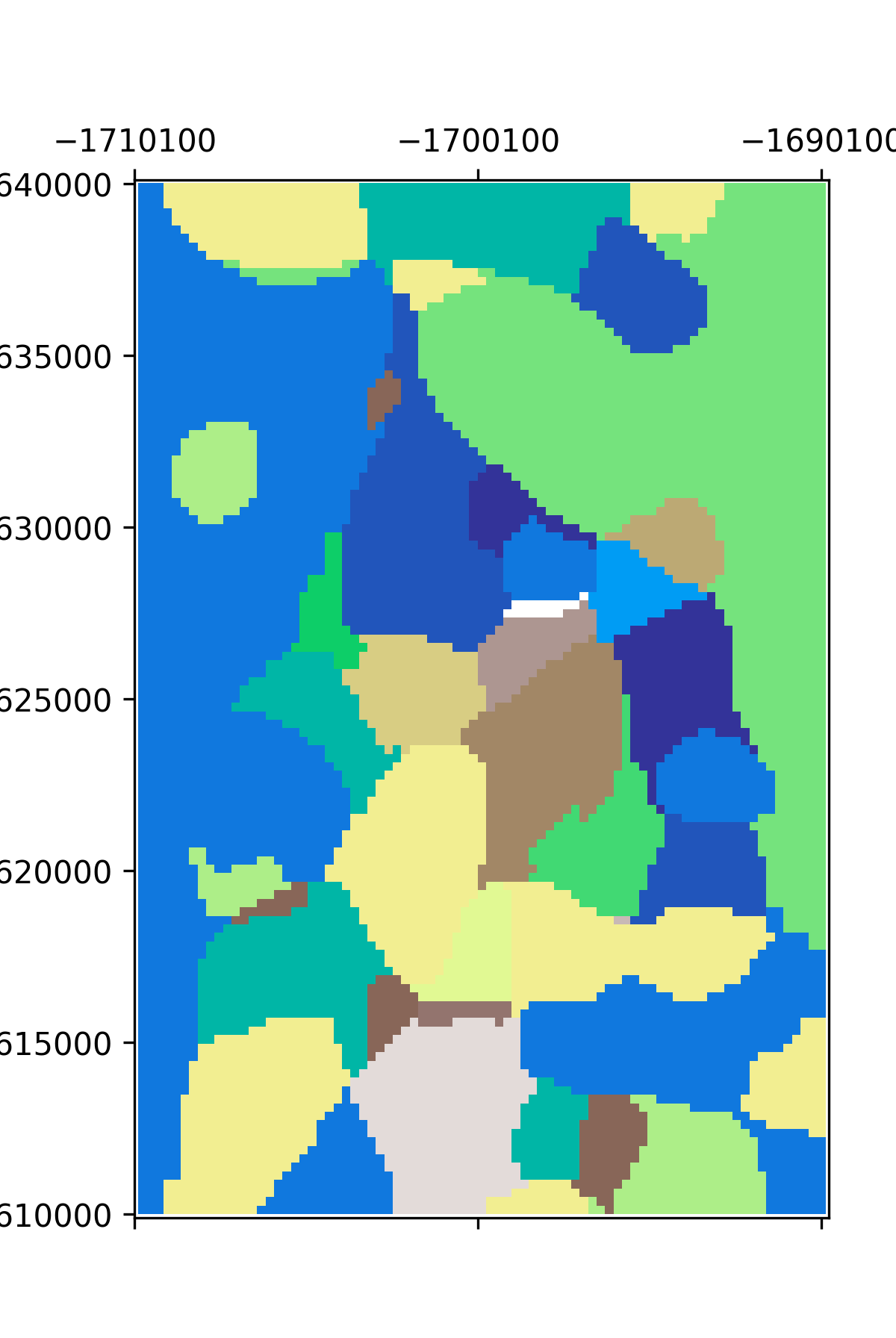}\vspace*{-0.35cm}
		\caption[]%
		{{\small  
		$\lambda_{min}$=1k 
		}}    
		\label{fig:theory1k}
	\end{subfigure}
	\hfill
	\begin{subfigure}[b]{0.136\textwidth}  
		\centering 
		\includegraphics[width=\textwidth]{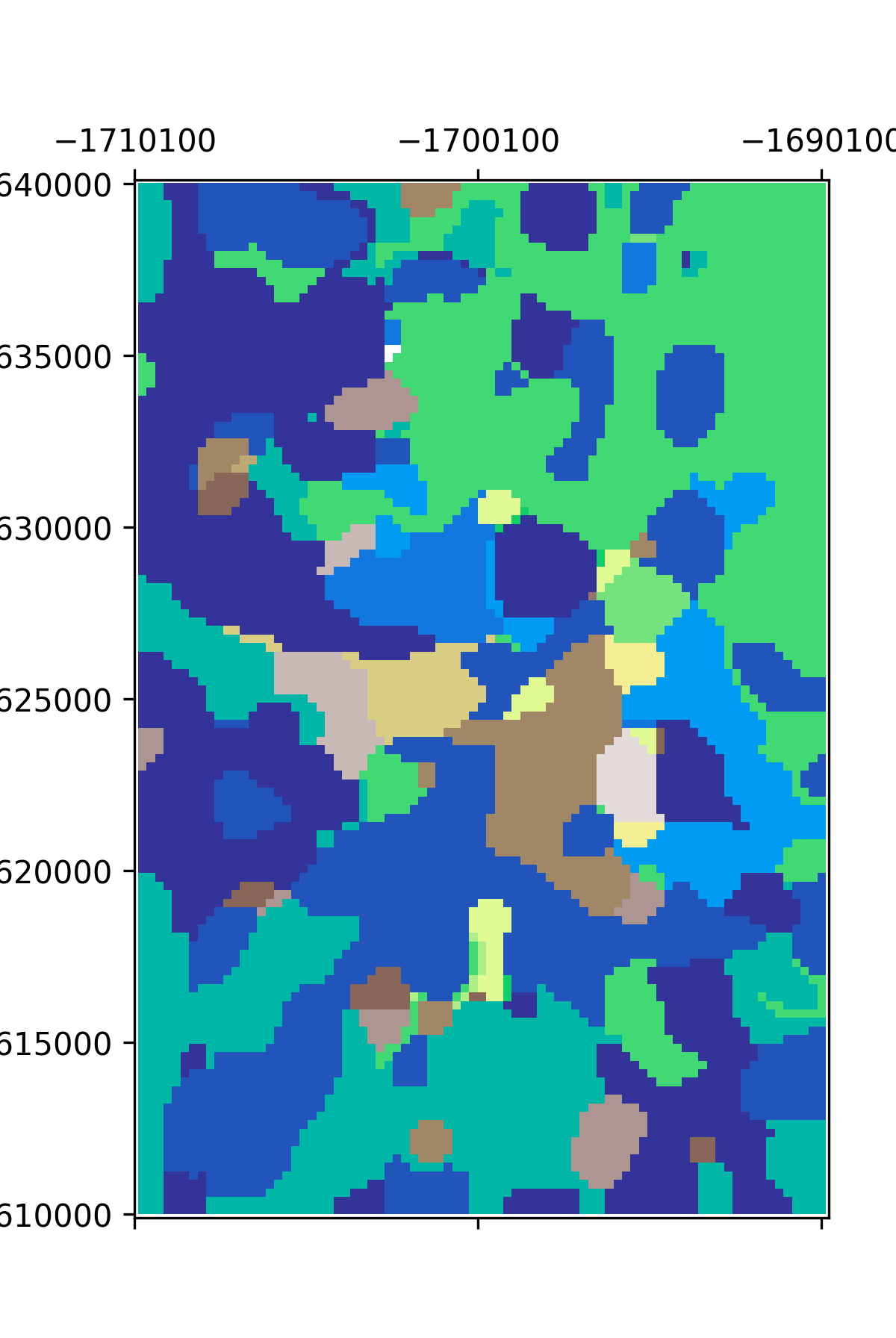}\vspace*{-0.35cm}
		\caption[]%
		{{\small  
		$\lambda_{min}$=500 
		}}    
		\label{fig:theory500}
	\end{subfigure}
	\hfill
	\begin{subfigure}[b]{0.142\textwidth}  
		\centering 
		\includegraphics[width=\textwidth]{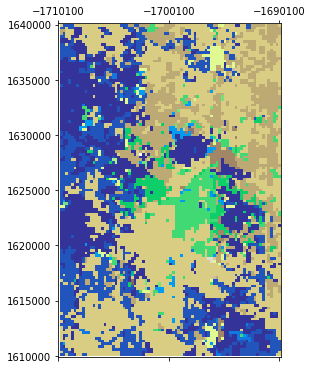}\vspace*{-0.1cm}
		\caption[]%
		{{\small 
		$\lambda_{min}$=50
		}}    
		\label{fig:theory40}
	\end{subfigure}
	\vspace*{-0.15cm}
	\caption{
	\small
	Embedding clustering  of (a) $\model{\naive}{loc}$; (b) $\model{tile}{loc}$ with the best cell size $c=500$; (c) $\model{\aodha}{loc}$ ($\numresnet=3,\numneuron=512$); (d) $\model{rbf}{loc}$ with the best $\sigma$ (1k) and 200 anchor points (red) and (e)(f)(h) $\model{theory}{loc}$ models with different $\lambda_{min}$, but fixed $\lambda_{max}=40k$ and $\nscale = 64$. 
	All models use 1 hidden ReLU layers of 512 neurons except $\model{\aodha}{cont}$. 
	} 
	\label{fig:loc_cluster}
	\vspace*{-0.15cm}
\end{figure*}
	

\begin{table}[]
	\caption{The evaluation results of different location models on the validation and test dataset.}
	\label{tab:loceval}
	\centering
	\small{ \begin{tabular}{c|c|ccc|cc}
		\toprule
		\multicolumn{1}{l|}{\multirow{2}{*}{}} & Train & \multicolumn{3}{c|}{Validation}           & \multicolumn{2}{c}{Testing}              \\ 
		\multicolumn{1}{l|}{}  & NLL    & NLL 
		& MRR        & HIT@5  & MRR      & HIT@5  \\ \hline
		$\model{random}{}$ &  & -
		& 0.052 (0.002) & 4.8 (0.5)        
		& 0.051 (0.002) & 5.0 (0.5)  \\ 
		$\model{\naive}{loc}$ 
		& 1.285 & 1.332 
		& 0.089 (0.001) & 10.6 (0.2)        
		& 0.090 (0.001) & 11.3 (0.2)  \\ 
		$\model{tile}{loc}$  
		($c$=500) 
		& 1.118 & 1.261 
		& 0.123 (0.001) & 16.8 (0.2)        
		& 0.120 (0.001) & 17.1 (0.3)  \\ 
		$\model{\aodha}{loc}$
		($\numresnet$=3,$\numneuron$=512) 
		& 1.222 & 1.288 
		& 0.112 (0.001) & 14.6 (0.1)        
		& 0.119 (0.001) & 15.8 (0.2)  \\ 
     

     	$\model{rbf}{loc}$ ($\sigma$=1k)  & 1.209& 1.279 
     	& 0.115 (0.001) & 15.2 (0.2)         
     	& 0.123 (0.001) & 16.8 (0.3) \\ 
     	 \hline
     	$\model{grid}{loc}$   ($\lambda_{min}$=50) & 1.156 & 1.258 
		& 0.128 (0.001) & 18.1 (0.3)       
		& 0.139 (0.001) & \textbf{20.0} (0.2) \\ 
		$\model{hexa}{loc}$  ($\lambda_{min}$=50)  & 1.230 & 1.297 
		& 0.107 (0.001) & 14.0 (0.2)        
		& 0.105 (0.001) & 14.5 (0.2) \\ 
		$\model{theorydiag}{loc}$  ($\lambda_{min}$=50)  & 1.277 & 1.324 
		& 0.094 (0.001) & 12.3 (0.3)         
		& 0.094 (0.002) & 11.2 (0.3)  \\  

		$theory$  ($\lambda_{min}$=1k)
		&   1.207 & 1.281 
		& 0.123 (0.002) & 16.3 (0.5)        
		& 0.121 (0.001) & 16.2 (0.1) \\

	$theory$  ($\lambda_{min}$=500) &   1.188&1.269 
		& 0.132 (0.001) & 17.6 (0.3)        
		& 0.129 (0.001) & 17.7 (0.2) \\ 

	$theory$  ($\lambda_{min}$=50)
	&   1.098&1.249 
		& \textbf{0.137} (0.002) & \textbf{19.4} (0.1)        
		& \textbf{0.144} (0.001)  & \textbf{20.0} (0.2) \\ 
		\bottomrule
	\end{tabular}}	\vspace*{-0.25cm}
\end{table}

We first study location modeling
with the \textbf{location decoder} $\locdec()$ in Section \ref{sec:contdec}.
%
We use a negative sample size of  $\negsize = 100$.
%
Table \ref{tab:loceval} shows the average metrics of different models 
with their best hyper-parameter setting
on the validation set. 
%
%
We can see that 
 $\model{\naive}{loc}$  and
 $\model{theorydiag}{loc}$ are less competitive, only beating the $\model{random}{loc}$ selection baseline.
%
Other methods with single scale representations -- including
 $\model{tile}{loc}$, $\model{\aodha}{loc}$, and $\model{rbf}{loc}$ -- perform better.
The best results come from various version of the grid cell models, which are capable of dealing with multi-scale representations.
 

In order to understand the reason for the superiority of
grid cell models
 we provide qualitative analysis of their representations.
We apply hierarchical clustering to the  location embeddings produced by studied models
using cosine distance as the distance metric (See Fig. \ref{fig:loc_cluster}).
%
we can see that when restricted to large grid sizes ($\lambda_{min} = 1k$), $\model{theory}{loc}$ has similar representation (Fig.~\ref{fig:rbfcluster}, \ref{fig:theory1k}, and Fig.~\ref{fig:locrbf}, \ref{fig:loctheory1k}) and performance compared to $\model{rbf}{loc}$ ($\sigma=1k$). However it is able to significantly outperform $\model{rbf}{loc}$ ($\sigma=1k$) (and $\model{tile}{loc}$ and $\model{\aodha}{loc}$) when small grid sizes ($\lambda_{min} = 500, 50$) are available. The relative improvements over $\model{rbf}{loc}$ ($\sigma=1k$) are -0.2\%, +0.6\%, +2.1\% MRR for $\lambda_{min}$=1k, 500, 50 respectively. 

\vsubsubsection{Multi-Scale Analysis of Location Modeling}
In order to show how our multi-scale location representation model will affect the prediction of POI types with different distribution patterns, we classify all 1,191 POI types into three groups based on radius $r$, which is derived from each POI types' renormalized Ripley's K curve (See Figure \ref{fig:ripley_log} for examples). 
It indicates the x axis value of the intersection between the curve and the line of $y = 3.0$. A lower $r$ indicates a more clustered distribution patterns. These three groups are listed below:
\begin{enumerate}
	\item Clustered ($r \leq 100m$): POI types with clustered distribution  patterns;
	\item Middle ($100m < r < 200m$): POI types with less extreme scales;
	\item Even ($r \geq 200m$): POI types with even distribution patterns.
\end{enumerate}

Table \ref{tab:locmol_stat} shows the performance ($MRR$) of $direct$, $tile$, $wrap$, $rbf$, and our $theory$ model on the test dataset of the location modeling task with respect to these three different POI distribution groups. The numbers in $()$ indicate the MRR difference betweeb a baseline and $theory$. \textit{\# POI} refers to total number of POI belong to each group\footnote{The reason why the sum of \textit{\# POI} of these three groups does not equal to the total number of POI is because one POI can have multiple types and they may belonging to different groups.}.
We can see that 
1) The two neural net approaches ($direct$ and $wrap$) have no scale related parameter and are not performing ideally across all scales, with $direct$ performs worse because of its simple single layer network.
2) The two approaches with built-in scale parameter ($tile$ and $rbf$) have to trade off the performance of different scales. Their best parameter settings lead to close performances to that of \modelname at the middle scale, while performing poorly in both clustered and regular groups.
These observation clearly shows that \textbf{all baselines can at most well handle distribution at one scale but show poor performances in other scales. In contrast, Space2Vec’s multi-scale representation can handle distributions at different scales.}

\begin{table}[t!]
	\caption{
	    Comparing performances in different POI groups.
		We classify all 1,191 POI types into three groups based on the radius $r$ of their root types, where their renormalized Ripley's K curve (See Figure \ref{fig:ripley_log}) reach 3.0: 
		1) Clustered ($r \leq 100m$): POI types with clustered distribution patterns; 
		2) Middle ($100m < r < 200m$): POI types with unclear distribution patterns; 
		3) Even ($r \geq 200m$): POI types with even distribution patterns. The MRR of $wrap$ and $theory$ on those three groups are shown. 
		The numbers in $()$ indicate the difference between the MRR of a baseline model and the MRR of $theory$ with respect to a specific group.
		\textit{\#POI} refers to the total number of POIs belonging to each group.
		\textit{Root Types} indicates the root categories of those POI types belong to each group.
	}
	\label{tab:locmol_stat}
	\centering
	\small{
		\begin{tabular}{c|c|c|c}
			\toprule
			\textbf{POI Groups} 
			&\textbf{Clustered}& \textbf{Middle} & \textbf{Even} \\
			 &  ($r \leq 100 m$)   &  ($100m < r < 200m$)   &  ($r \geq 200 m$)  \\ \hline
			$direct$   
			& 0.080 (-0.047) & 0.108 (-0.030)    & 0.084 (-0.047)   \\
			$wrap$   
			& 0.106 (-0.021) & 0.126 (-0.012)    & 0.122 (-0.009)   \\  \hline
			$tile$   
			& 0.108 (-0.019) & 0.135 (-0.003)    & 0.111 (-0.020)   \\
			$rbf$   
			& 0.112 (-0.015) & 0.136 (-0.002)    & 0.119 (-0.012)   \\ \hline
			$theory$
			& 0.127 (-)  & 0.138  (-)   & 0.131 (-)   \\ \hline
			{\# POI}       & 16,016  & 7,443   & 3,915   \\ \hline
			{Root Types}    & \begin{tabular}[c]{@{}l@{}}
			Restaurants; Shopping; Food; \\
			Nightlife;  Automotive; Active \\
			 Life; Arts \& Entertainment; \\
			Financial Services\end{tabular} 
			& 
			\begin{tabular}[c]{@{}l@{}}
			Beauty \& Spas; Health \& Medical;\\
			Local Services; Hotels \& Travel;\\
			Professional Services;\\
			Public Services  \& Government
			\end{tabular} 
			& 
			\begin{tabular}[c]{@{}l@{}}
			Home Services; \\
			Event Planning \\ 
			\& Services;\\
			 Pets; Education
			 \end{tabular} 
			 \\ \bottomrule
		\end{tabular}
	}	\vspace*{-0.25cm}
\end{table}

\vsubsubsection{Spatial Context Modeling Evaluation}  \label{sec:conteval}

\begin{table}[b!]
	\caption{The evaluation results of different spatial context models on the validation and test dataset. 
	All encoders contains a 1 hidden layer FFN.
	All grid cell encoders set $\lambda_{min}$=10, $\lambda_{max}$=10k.
	}
	\label{tab:conteval}
	\centering
	\small{ 
	\begin{tabular}{c|c|ccc|cc}
		\toprule
		 & Train & \multicolumn{3}{c}{Validation} &    
		 \multicolumn{2}{c}{Testing}              \\
		$\modelname$ & NLL & NLL  & MRR & HIT@5   & MRR     & HIT@5       \\ \hline
		$\model{none}{cont}$         
		& 1.163 & 1.297 & 0.159 (0.002) & 22.4 (0.5) 
		& 0.167 (0.006) & 23.4 (0.7) \\ 
		$\model{\naive}{cont}$ 
		& 1.151 & 1.282  & 0.170 (0.002) & 24.6 (0.4) 
		&  0.175 (0.003) & 24.7 (0.5) \\ 
        
		$\model{polar}{cont}$
		& 1.157 & 1.283 & 0.176 (0.004) & 25.4 (0.4) 
		& 0.178 (0.006) & 24.9 (0.1) \\ 
		$\model{tile}{cont}$
		$(c=50)$        
		& 1.163 & 1.298  & 0.173 (0.004) & 24.0 (0.6) 
		&  0.173 (0.001) & 23.4 (0.1) \\ 

		$\model{polar\_tile}{cont} (\freq=64)$        
		& 1.161 & 1.282  & 0.173 (0.003) & 25.0 (0.1) 
		&  0.177 (0.001) & 24.5 (0.3) \\ 

		$\model{\aodha}{cont}$ 
		($\numresnet$=2,$\numneuron$=512)      
		& 1.167 & 1.291  & 0.159 (0.001) & 23.0 (0.1) 
		&  0.170 (0.001) & 23.9 (0.2) \\ 

		$\model{rbf}{cont}$ $(\sigma=50)$
		&  1.160 & 1.281  & \textbf{0.179} (0.002) & 25.2 (0.6) 
		&  0.172 (0.001) &  25.0 (0.1) \\ 
		
		$\model{scaled\_rbf}{cont}$ ($\sigma$=40,$\beta$=0.1) 
		&  1.150 & 1.272 & 0.177 (0.002) & \textbf{25.7} (0.1) 
		&  0.181 (0.001) &  25.3 (0.1) \\ 
\hline

		$\model{grid}{cont}$($\lambda_{min}$=10)
		& 1.172 & 1.285 & 0.178 (0.004) & 24.9 (0.5) 
		& 0.181 (0.001) & 25.1 (0.3) \\ 
		$\model{hexa}{cont}$ ($\lambda_{min}$=10)
		& 1.156 & 1.289  & 0.173 (0.002) & 24.0 (0.2) 
		& 0.183 (0.002) & 25.3 (0.2) \\ 
		$\model{theorydiag}{cont}$ $(\lambda_{min}=10)$
		& 1.156 & 1.287 &  0.168 (0.001) & 24.1 (0.4) 
		& 0.174 (0.005) & 24.9 (0.1) \\ 
		$\model{theory}{cont}$($\lambda_{min}$=200)
		& 1.168 & 1.295 & 0.159 (0.001) & 23.1 (0.2) 
		& 0.170 (0.001) &  23.2 (0.2) \\
		$\model{theory}{cont}$($\lambda_{min}$=50)
		& 1.157 & 1.275 & 0.171 (0.001) & 24.2 (0.3) 
		& 0.173 (0.001) &  24.8 (0.4) \\
		$\model{theory}{cont}$($\lambda_{min}$=10)
		& 1.158 & 1.280 & 0.177 (0.003) & 25.2 (0.3) 
		& \textbf{0.185} (0.002) &  \textbf{25.7} (0.3) \\
		\bottomrule
	\end{tabular} }	
\end{table}

Next, we evaluate the 
 \textbf{spatial context decoder} $\contdec()$ in Sec. \ref{sec:contdec}. 
We use the same evaluation set up as location modeling. 
The context points are obtained by querying the $\neisize$-th nearest points  using PostGIS ($\neisize = 10$). 
As for validation and test datasets, we make sure the center points are all unknown during the training phase. 
Table \ref{tab:conteval} shows the evaluation results of different models
for spatial context modeling. 
The baseline approaches 
($\model{\naive}{cont}$, $\model{tile}{cont}$,  $\model{\aodha}{cont}$, $\model{rbf}{cont}$) generally perform poorly in context modeling.
We designed specialized version of these approaches ($\model{polar}{cont}$,  $\model{polar\_tile}{cont}$, $\model{scaled\_rbf}{cont}$) with polar coordinates, which lead to significantly improvements.
Note that these are  models proposed by us specialized for context modeling and therefore are less general than the grid cell approaches.
Nevertheless the grid cell approaches
are able to perform better than the specialized approaches  on the test dataset while have competitive performance on validation dataset.
See Appendix~\ref{sec:apdx:context} for the visualization of context models.
%
Actually the gains are small for all baseline approaches also. The reason is that we expect location encoding to be less important when context information is accessible. Similarly as discussed in \citep{gao2018learning}, it is when there is a lack of visual clues that the grid cells of animals are  the most helpful for their navigation. 

\begin{figure*}[t!]
	\centering \tiny
	\vspace*{-0.2cm}
	\begin{subfigure}[b]{0.16\textwidth}  
		\centering 
		\includegraphics[width=\textwidth]{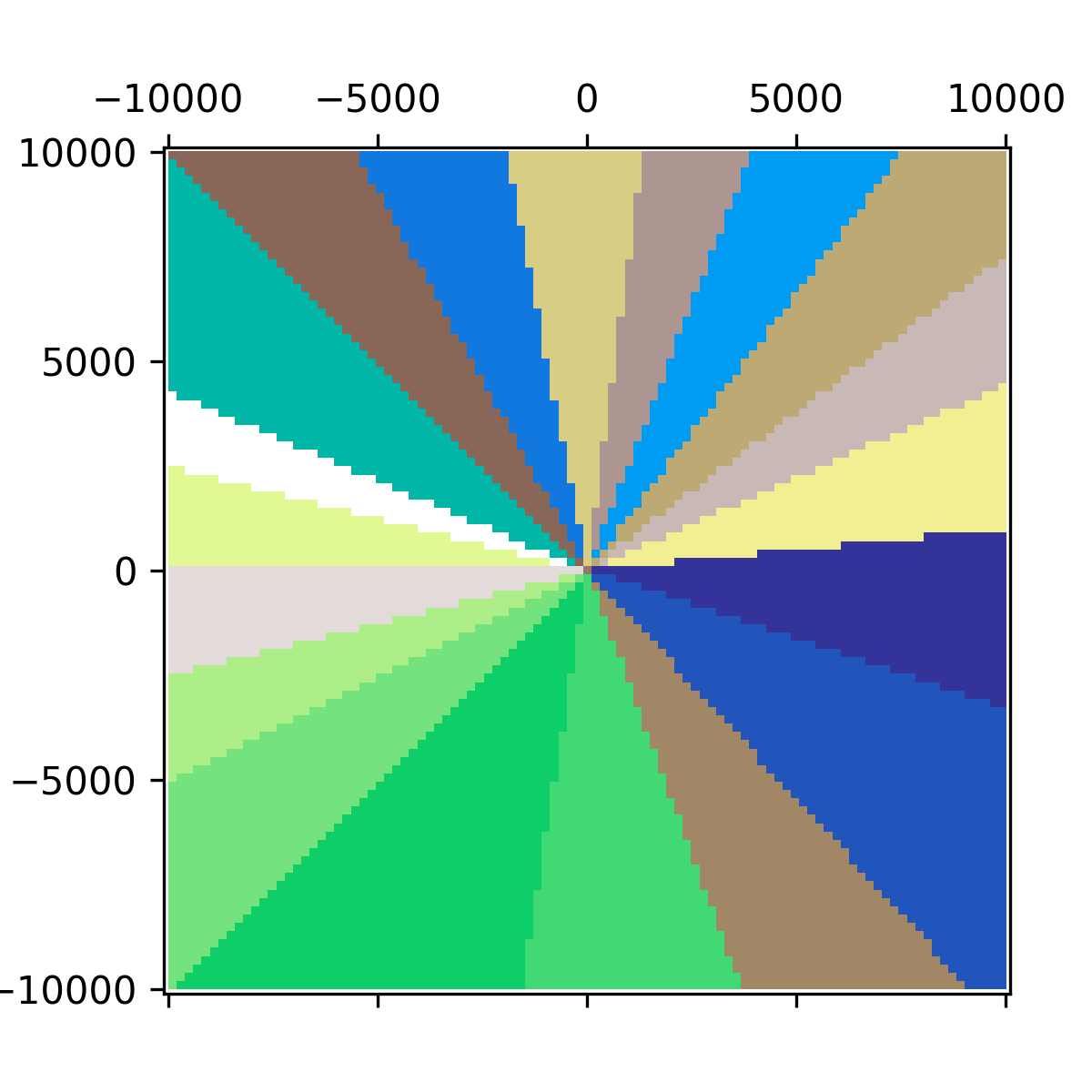}\vspace*{-0.2cm}
		\caption[]%
		{{\small 
		$\model{\naive}{cont}$
		}}    
		\label{fig:cont_naive}
	\end{subfigure}
	\hfill
	\begin{subfigure}[b]{0.165\textwidth}  
		\centering 
		\includegraphics[width=\textwidth]{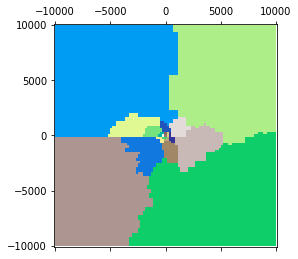}\vspace*{-0.1cm}
		\caption[]%
		{{\small 
		$\model{polar}{cont}$
		}}    
		\label{fig:cont_polar}
	\end{subfigure}
	\hfill
	\begin{subfigure}[b]{0.16\textwidth}  
		\centering 
		\includegraphics[width=\textwidth]{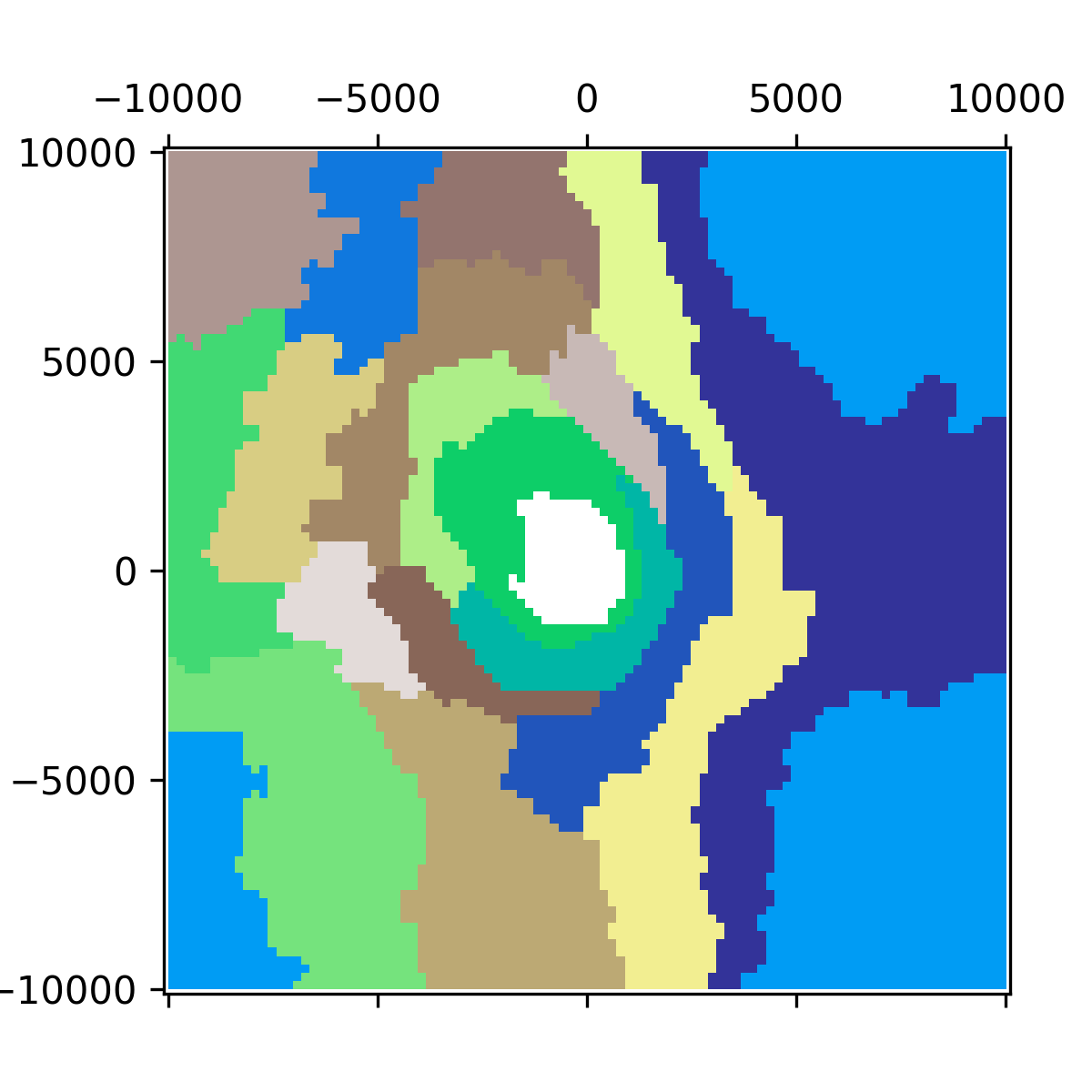}\vspace*{-0.2cm}
		\caption[]%
		{{\small 
		$\model{\aodha}{cont}$ 
		}}    
		\label{fig:cont_aodha}
	\end{subfigure}
	\hfill
	\begin{subfigure}[b]{0.16\textwidth}  
		\centering 
		\includegraphics[width=\textwidth]{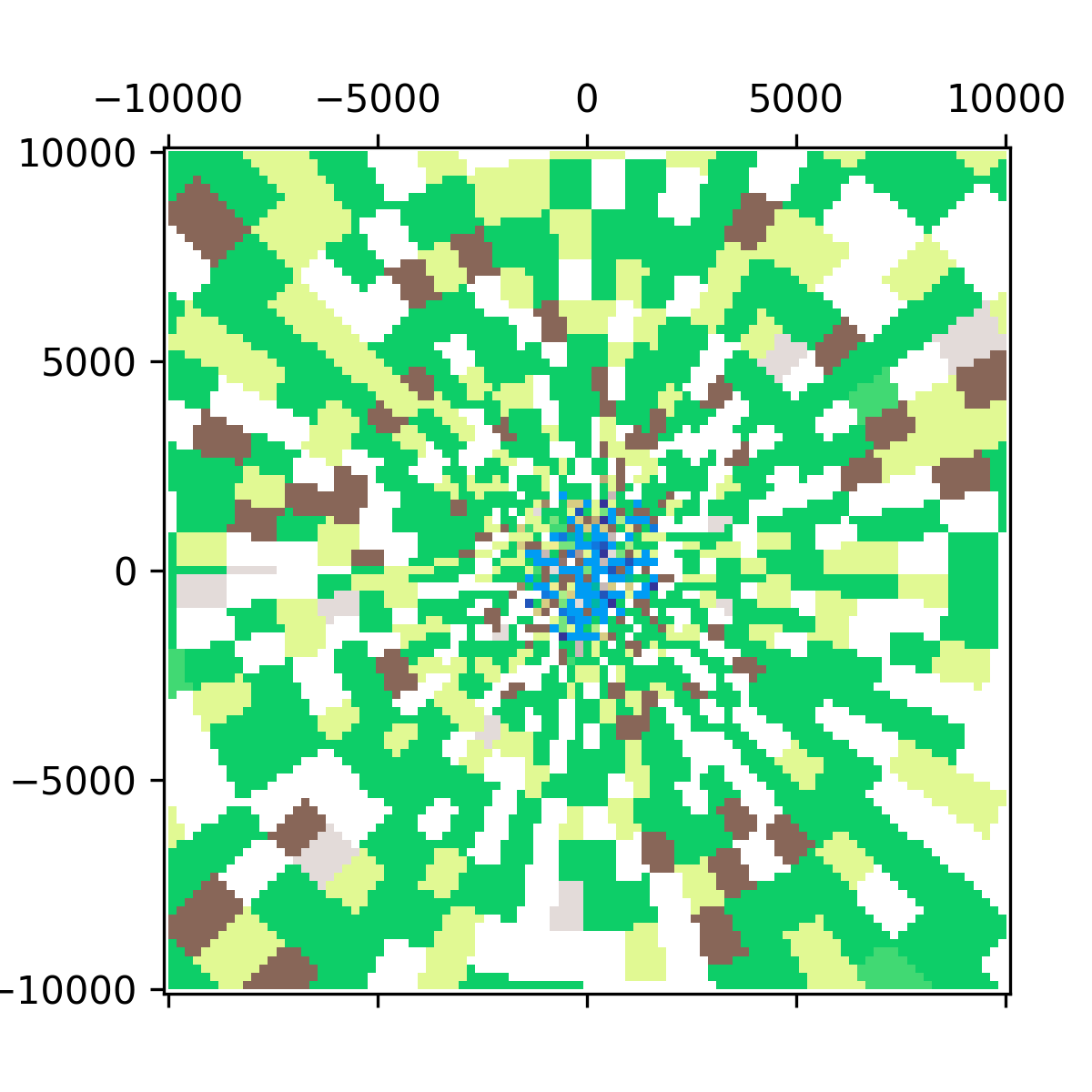}\vspace*{-0.15cm}
		\caption[]%
		{{\small  
		$\model{polar\_tile}{cont}$ 
		}}    
		\label{fig:cont_polar_tile}
	\end{subfigure}
	\hfill
    \begin{subfigure}[b]{0.165\textwidth}  
		\centering 
		\includegraphics[width=\textwidth]{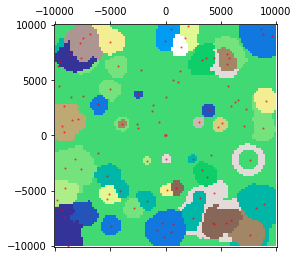}\vspace*{-0.03cm}
		\caption[]%
		{{\small  
		$\model{scaled\_rbf}{cont}$ 
		}}    
		\label{fig:cont_rbf_01_}
	\end{subfigure}
	\hfill
	\begin{subfigure}[b]{0.16\textwidth}  
		\centering 
		\includegraphics[width=\textwidth]{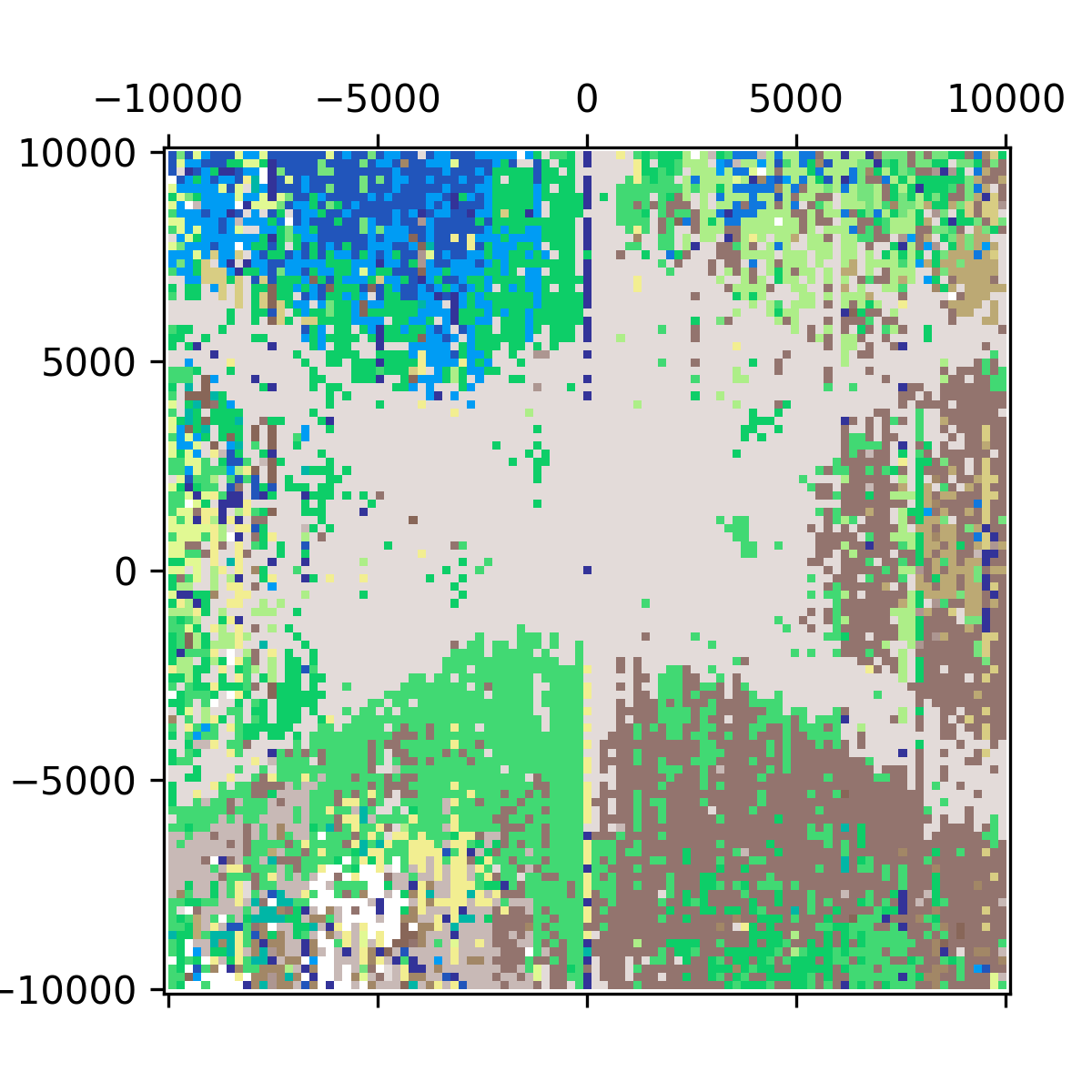}\vspace*{-0.2cm}
		\caption[]%
		{{\small 
		$\model{theory}{cont}$ 
		}}    
		\label{fig:cont_theory_10_}
	\end{subfigure}
	\begin{subfigure}[b]{0.155\textwidth}  
		\centering 
		\includegraphics[width=\textwidth]{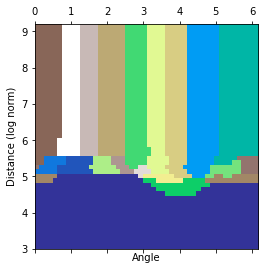}\vspace*{-0.2cm}
		\caption[]%
		{{\small 
		$\model{\naive}{cont}$
		}}    
		\label{fig:cont_naive_polar}
	\end{subfigure}
	\hfill
	\begin{subfigure}[b]{0.155\textwidth}  
		\centering 
		\includegraphics[width=\textwidth]{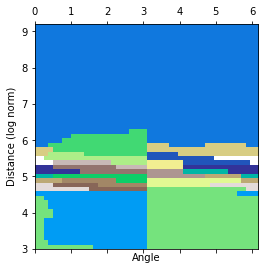}\vspace*{-0.2cm}
		\caption[]%
		{{\small 
		$\model{polar}{cont}$
		}}    
		\label{fig:cont_polar_polar}
	\end{subfigure}
	\hfill
	\begin{subfigure}[b]{0.16\textwidth}  
		\centering 
		\includegraphics[width=\textwidth]{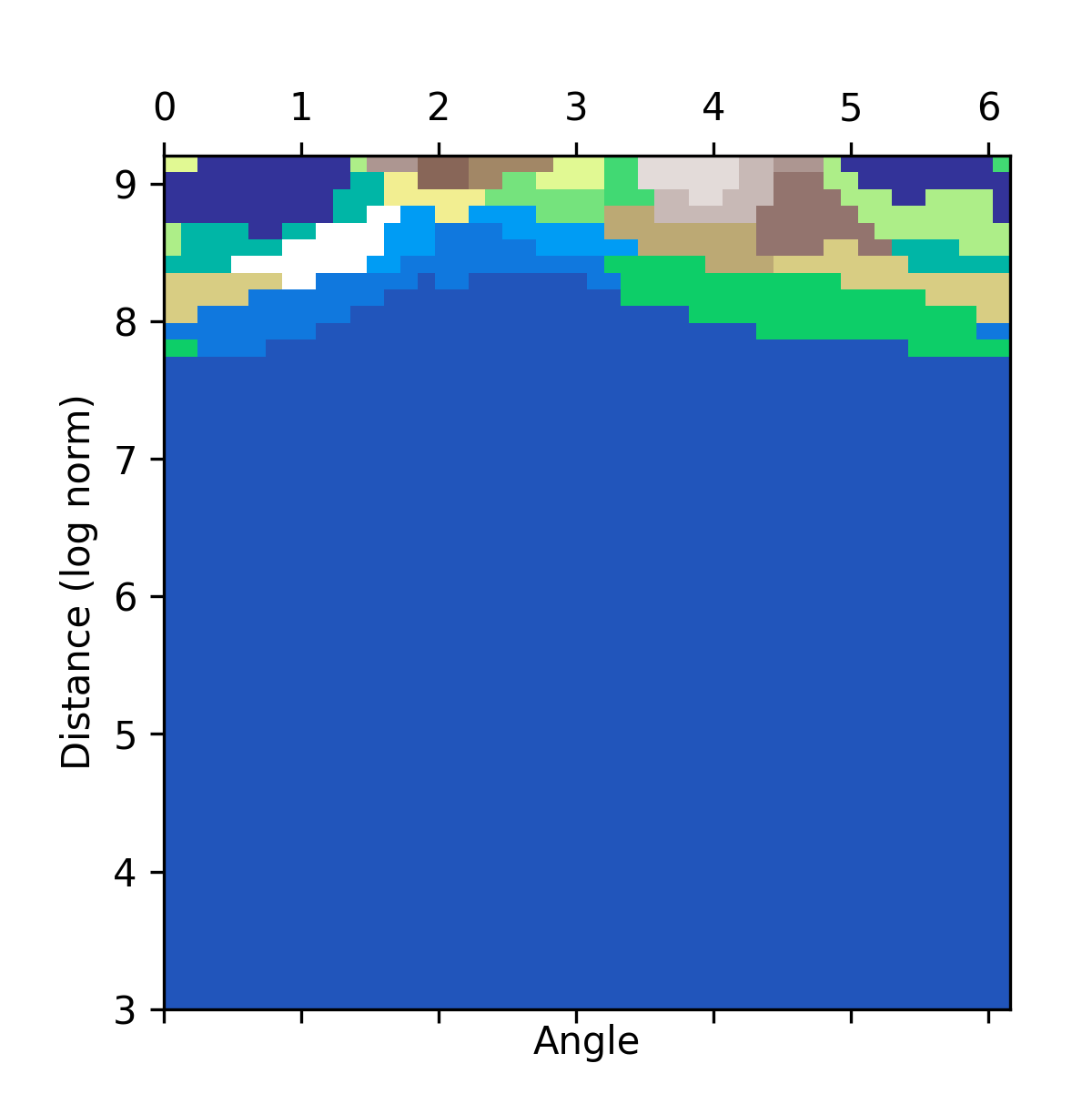}\vspace*{-0.2cm}
		\caption[]%
		{{\small 
		$\model{\aodha}{cont}$ 
		}}    
		\label{fig:cont_aodha_polar}
	\end{subfigure}
	\hfill
	\begin{subfigure}[b]{0.16\textwidth}  
		\centering 
		\includegraphics[width=\textwidth]{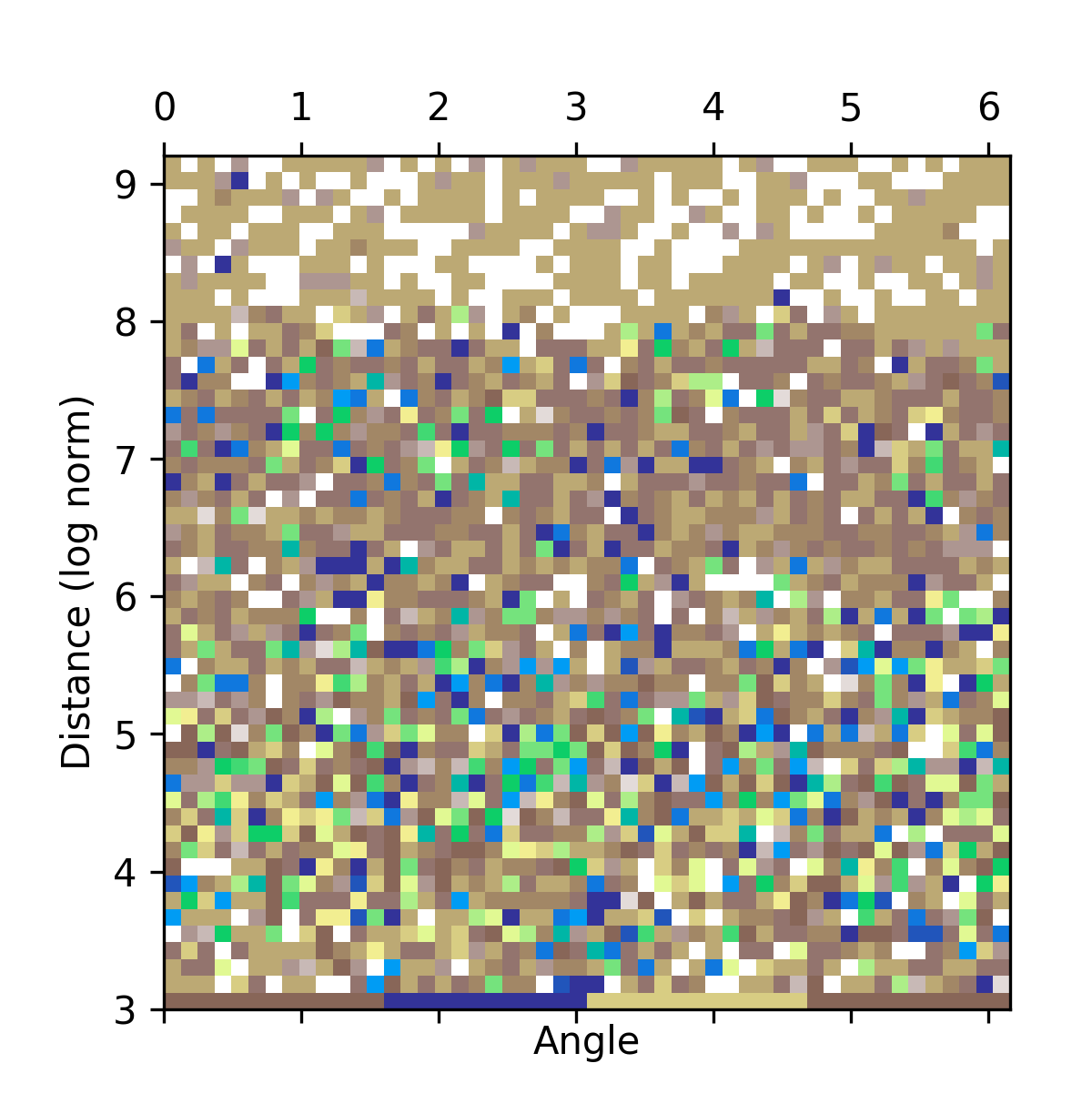}\vspace*{-0.2cm}
		\caption[]%
		{{\small  
		$\model{polar\_tile}{cont}$ 
		}}    
		\label{fig:cont_polar_tile_polar}
	\end{subfigure}
	\hfill
	\begin{subfigure}[b]{0.145\textwidth}  
		\centering 
		\includegraphics[width=\textwidth]{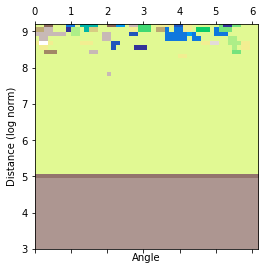}\vspace*{-0.1cm}
		\caption[]%
		{{\small  
		$\model{scaled\_rbf}{cont}$ 
		}}    
		\label{fig:cont_rbf_01_polar_}
	\end{subfigure}
	\hfill
	\begin{subfigure}[b]{0.16\textwidth}  
		\centering 
		\includegraphics[width=\textwidth]{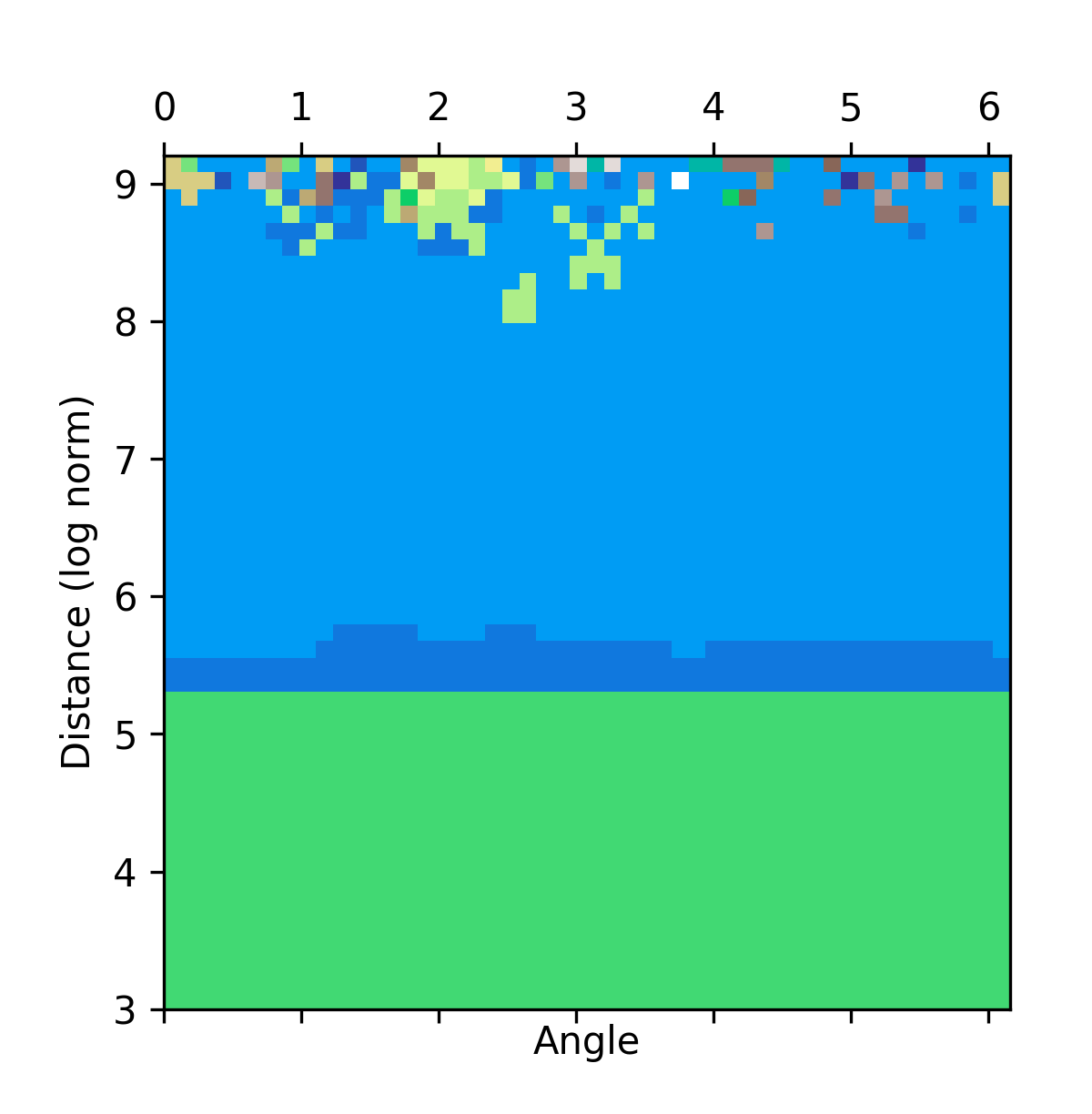}\vspace*{-0.2cm}
		\caption[]%
		{{\small 
		$\model{theory}{cont}$ 
		}}    
		\label{fig:cont_theory_10_polar_}
	\end{subfigure}
	\caption{
	Embedding clustering in the original space of 
	(a) $\model{\naive}{cont}$; 
	(b) $\model{polar}{cont}$; 
	(c) $\model{\aodha}{cont}$, $\numresnet$=2,$\numneuron$=512; 
	(d) $\model{polar\_tile}{cont}$, $\freq$ = 64, 
	(e) $\model{scaled\_rbf}{cont}$, $\sigma$ = 40, $\beta$=0.1; and
	(f) $\model{theory}{cont}$, $\lambda_{min}=10$, $\lambda_{max}=10k$, $\nscale = 64$. 
	(g)(h)(i)(j)(k)(l) are the  clustering results of the same models in the polar-distance space using $\log(\parallel\Delta\bx_{ij}\parallel + 1)$. All models use 1 hidden ReLU (except $\model{\aodha}{cont}$) layers of 512 neurons. 
	Most models except  $\model{\aodha}{cont}$ can capture a shift when distance is around $e^5-1 \approx 150$ meters.
	} 
	\label{fig:poicl}
	\vspace*{-0.15cm}
\end{figure*}

Figure~\ref{fig:poicl} shows the location embedding clustering results in both 
Cartesian and polar  coordinate systems.
We can see that  $\model{\naive}{cont}$ (Fig. \ref{fig:cont_naive}, \ref{fig:cont_naive_polar}) only captures  the distance information when the context POI is very  close ($log(\parallel\Delta\bx_{ij}\parallel + 1) \leq 5$) while in the farther spatial context it purely models the direction information. 
$\model{polar}{cont}$ (Fig. \ref{fig:cont_polar}, \ref{fig:cont_polar_polar}) has the similar behaviors but captures the distance information in a more fine-grained manner. 
$\model{\aodha}{cont}$ (Fig. \ref{fig:cont_aodha}, \ref{fig:cont_aodha_polar}) mainly focuses on differentiating relative positions in farther spatial context $\cont$ which might explain its lower performance\footnote{Note that $\model{\aodha}{cont}$ is original proposed by \citet{mac2019presence} for location modelling, not spatial context modelling. This results indicates $\model{\aodha}{cont}$  is not good at this task.}.
$\model{polar\_tile}{cont}$ (Fig. \ref{fig:cont_polar_tile})
mostly responds to distance information. 
Interestingly, $\model{scaled\_rbf}{cont}$ and $\model{theory}{cont}$  have similar representations in the polar coordinate system (Fig. \ref{fig:cont_rbf_01_polar_}, \ref{fig:cont_theory_10_polar_}) and similar performance (Table \ref{tab:conteval}). 
While $\model{scaled\_rbf}{cont}$  captures the gradually decreased distance effect with a scaled kernel size which becomes larger in farther distance, $\model{theory}{cont}$ achieves this by integrating representations of different scales. 

\subsection{Fine-Grained Image Classification Tasks}
To demonstrate the generalizability of  \modelname
for space representation
we utilized the proposed point space encoder $\enc^{(x)}()$ model in a well-known computer vision task: \textit{fine-grained image classification}.
As we discussed in Section \ref{sec:related}, many studies \citep{berg2014birdsnap,chu2019geo,mac2019presence} have shown that geographic prior information - where (and when) the image is taken - is very important additional information for the fine-grained image classification task and can substantially improve the model performance. 
For example, the appearance information is usually not sufficient to differentiate two visually similar species. 
In this case, the geographic prior becomes much more important because these two species may have very different spatial prior distributions such as the example of European Toads and Spiny Toads in Figure 1 of \citet{mac2019presence}.

We adopt the task setup of \citet{mac2019presence}. During training we have a set of tuples $D = \{(I_{i}, \bx_{i}, y_{i}, p_{i}) \mid i = 1,...,N\}$ where $I_{i}$ indicates an image, $y_{i} \in \{1,2,...,C\}$ is the corresponding class label (species category), $\bx_{i} = [longitude_{i}, latitude_{i}]$ is the geographic coordinates where the image was taken, and $p_{i}$ is the id of the photographer who took this image. At training time, a location encoder is trained to capture the spatial prior information $P(y \mid \bx)$. At inference time, $p_{i}$ information is not available and the final image classification prediction is calculated based on the combination of two models: 1) the trained location encoder which captures the spatial priors $P(y \mid \bx)$ and 2) the pretrained image classification model, InceptionV3 network \citep{szegedy2016rethinking}, which captures $P(y \mid I)$. Bayesian theory has been used to derive the joint distribution $P(y \mid I, \bx)$. See \citet{mac2019presence} for detail explanation as well as the loss function.
Note that 
while \modelname outperforms specialized density estimation methods such as Adaptive Kernel \citep{berg2014birdsnap}, it would be interesting  to explore   early fusion \modelname's representations with the image module.

We use two versions of our point space encoder $\enc^{(x)}()$ model ($grid$, $theory$) as the location encoder to capture the spatial prior information $P(y \mid \bx)$. The evaluation results of our models as well as multiple baselines are shown in Table \ref{tab:imgcls_eval}. We can see that \textbf{both $grid$, $theory$ outperform previous models as well as that of \citet{mac2019presence}   on two fine-grained image classification datasets  with significant sizes: BirdSnap$\dagger$, NABirds$\dagger$}. $theory$ shows superiority over $grid$ on NABirds$\dagger$ while fail to outperform $grid$ on BirdSnap$\dagger$. Note that we only pick baseline models which capture spatial-only prior and drop models which additionally consider time information. Both $grid$ and $theory$ use 1 hidden ReLU layers of 512 neurons for $\mathbf{NN}()$ and they have the same hyperparameters: $\lambda_{min}$=0.0001, $\lambda_{max}$=360, $\nscale$ = 64. Like \citet{mac2019presence}, the location embedding size $\pedim$ is 1024 and we train the location encoder for 30 epochs. Our implementation is based on the original code\footnote{\url{https://github.com/macaodha/geo_prior/}} of \citet{mac2019presence} for both model training and evaluation phase.

\begin{table}[t!]
	\caption{Fine-grained image classification results on two datasets: BirdSnap$\dagger$ and NABirds$\dagger$. The classification accuracy is calculated by combining image classification predictions $P(y \mid I)$ with different spatial priors $P(y \mid \bx)$. The $grid$ and $theory$ model use 1 hidden ReLU layers of 512 neurons. The evaluation results of the baseline models are  from Table 1 of \citet{mac2019presence}. 
	}
	\label{tab:imgcls_eval}
	\centering
	\small{
		\begin{tabular}{l|l|l}
			\toprule
			& BirdSnap$\dagger$      & NABirds$\dagger$       \\ \hline
			No Prior (i.e. uniform)     & 70.07          & 76.08          \\ \hline
			Nearest Neighbor (num)      & 77.76          & 79.99          \\
			Nearest Neighbor (spatial)  & 77.98          & 80.79          \\ \hline
			Adaptive Kernel \citep{berg2014birdsnap}     & 78.65          & 81.11          \\
			$tile$
			\citep{tang2015improving} (location only)         & 77.19          & 79.58          \\
			$\aodha$ \citep{mac2019presence}  (location only) & 78.65          & 81.15          \\ 
			$rbf$ ($\sigma$=1k)     & 78.56 & 81.13          \\
			$grid$ ($\lambda_{min}$=0.0001, $\lambda_{max}$=360, $\nscale$ = 64)  & \textbf{79.44} & 81.28          \\
			$theory$ ($\lambda_{min}$=0.0001, $\lambda_{max}$=360, $\nscale$ = 64) & 79.35          & \textbf{81.59} \\
			\bottomrule
		\end{tabular}
	}	\vspace*{-0.25cm}
\end{table}

\vsection{Conclusion}   

We introduced an encoder-decoder framework  as a general-purpose representation model for space inspired by biological grid cells'  multi-scale periodic representations.
The model is an inductive learning model and can be trained in an unsupervised manner. 
We conduct two experiments on POI type prediction based on 1) POI locations and 2) nearby POIs. The evaluation results demonstrate the effectiveness of our model. 
Our analysis reveals that it is the ability to integrate representations of different scales that makes the grid cell models outperform other baselines on these two tasks.
In the future, we hope to incorporate the presented framework to more complex GIS tasks 
such as social network analysis, and sea surface temperature prediction.


\subsubsection*{Acknowledgments}
The presented work is partially funded by the NSF award 1936677 C-Accel Pilot - Track A1 (Open Knowledge Network): \textit{Spatially-Explicit Models, Methods, And Services For Open Knowledge Networks}, Esri Inc., and Microsoft AI for Earth Grant: \textit{Deep Species Spatio-temporal Distribution Modeling for Biodiversity Hotspot Prediction}. We thank Dr. Ruiqi Gao for discussions about grid cells, Dr. Wenyun Zuo for discussion about species potential distribution prediction and Dr. Yingjie Hu for his suggestions about the introduction section. 

\bibliographystyle{iclr2020_conference}
\bibliography{reference}

\appendix
\newpage
\section{Appendix}

\subsection{ Baselines }\label{sec:baselines}

To help understand the mechanism of distributed space representation we compare multiple ways of encoding spatial information.
Different models use different point space encoder $\enc^{(x)}()$ to encode either location $\bx_i$ (for location modeling $\loc$) or the displacement between the center point and one context point $\Delta\bx_{ij} = \bx_{i}-\bx_{ij}$ (for spatial context modeling $\cont$)\footnote{
We will use meter as the unit of $\lambda_{min}, \lambda_{max}, \sigma, c$.
}.
 \begin{itemize}
 \item $random$
 shuffles the order of the correct POI and $\negsize$ negative samples randomly as the predicted ranking. This shows the lower bound of each metrics.

 \item 	${\naive}$
directly encode location $\bx_i$ 
(or $\Delta\bx_{ij}$ for $\cont$) into a location embedding $\peemb{i}$ (or $\mathbf{e}[\Delta\bx_{ij}]$) using a feed-forward neural networks  (FFNs)\footnote{we first normalizes $\bx$ (or $\Delta\bx$) into the range $[-1,1]$}, denoted as $\enc_{\naive}^{(x)}(\bx)$ without decomposing coordinates into a multi-scale periodic representation. 
This is essentially the GPS encoding method used by \citet{chu2019geo}. 
Note that \citet{chu2019geo} is not open sourced and we end up implementing the model architecture ourselves. 

\item $tile$
divides the study area $A_{loc}$ (for $\loc$) or the range of spatial context defined by $\lambda_{max}$, $A_{cont}$, (for $\cont$) into grids with equal grid sizes $c$. Each grid has an embedding to be used as the encoding for every location $\bx_i$ or displacement $\Delta\bx_{ij}$ fall into this grid. 
This is a common practice  by many previous work when  dealing with coordinate data \citep{berg2014birdsnap,adams2015frankenplace,tang2015improving}.

\item $\aodha$
is a location encoder model recently introduced by \citet{mac2019presence}. It first normalizes $\bx$ (or $\Delta\bx$) into the range $[-1,1]$ and uses a coordinate wrap mechanism $[\sin(\pi \bx^{[l]});\cos(\pi \bx^{[l]})]$ to convert each dimension of $\bx$ into 2 numbers. This is then passed through an initial fully connected layer, followed by a series of $\numresnet$ residual blocks, each consisting of two fully connected layers ($\numneuron$ hidden neurons) with a dropout layer in between. We adopt the official code of \citet{mac2019presence}\footnote{\url{http://www.vision.caltech.edu/~macaodha/projects/geopriors/}} for this implementation.

\item ${rbf}$ 
randomly samples $M$ points from the training dataset as RBF anchor points \{$\bx^{anchor}_{m}, m=1...M$\} (or samples $M$ $\Delta\bx^{anchor}_{m}$ from $A_{cont}$ for $\cont$) \footnote{these anchor points are fixed in both $\loc$ and $\cont$.}, and use gaussian kernels $\exp{\big(-\dfrac{\parallel \bx_i  - \bx^{anchor}_{m} \parallel^2}{2\sigma^2}\big)}$ (or $\exp{\big(-\dfrac{\parallel \Delta\bx_{ij}  - \Delta\bx^{anchor}_{m} \parallel^2}{2\sigma^2}\big)}$ for $\cont$) on each anchor points, where $\sigma$ is the kernel size. 
Each point $p_i$ has a $M$-dimension RBF feature vector which is fed into a FNN to obtain the spatial embedding. This is a strong baseline for representing floating number features in machine learning models.

\item $grid$ 
 as described in Section \ref{sec:peenc}  inspired by the position encoding in Transformer \citep{vaswani2017attention}.
\item ${hexa}$ Same as  $grid$ but 
use $sin(\theta)$, $sin(\theta+2\pi/3)$, and $sin(\theta+4\pi/3)$ 
in $PE^{(g)}_{s,l}(\bx)$.
\item  $theory$ as described in Section \ref{sec:peenc}, uses the theoretical models  \citep{gao2018learning} as the first layer of $\enc_{theory}^{(x)}(\bx)$ or $\enc_{theory}^{(x)}(\Delta\bx_{ij})$.
\item ${theorydiag}$ further constrains $\pemlp()$ as a block diagonal matrix, with each scale as a block. 
\end{itemize}

We also have the following baselines which are specific to the spatial context modeling task.
\begin{itemize}
 \item $none$ the decoder
$\contdec()$  does not consider the spatial relationship between the center point and context points but only the co-locate patterns such as Place2Vec \citep{yan2017itdl}. That means we drop the $\mathbf{e}[\Delta\bx_{ij}]$ from the attention mechanism in Equ. \ref{equ:atten_gc} and \ref{equ:atten_init}.

\item ${polar}$ first converts the displacement $\Delta\bx_{ij}$ into polar coordinates $(r, \theta)$ centered at the center point 
where $r = log(\parallel\Delta\bx_{ij}\parallel + 1)$
. Then it uses  $[r, \theta]$ as the input for a  FFN to obtain the spatial relationship embedding in Equ. \ref{equ:atten_gc}. 
We find out that it 
has a significant performance improvement over the variation with $r = \parallel\Delta\bx_{ij}\parallel$.

\item $polar\_tile$
is a modified version of $tile$ but the grids are extracted from polar coordinates $(r, \theta)$ centered at the center point where $r = log(\parallel\Delta\bx_{ij}\parallel + 1)$. Instead of using grid size $c$, we use the number of grids along $\theta$ (or $r$) axis, $F$, as the only hyperparameter. Similarly, We find that  $r = log(\parallel\Delta\bx_{ij}\parallel + 1)$ outperform  $r = \parallel\Delta\bx_{ij}\parallel$ significantly.

\item ${scaled\_rbf}$ 
is a modified version of ${rbf}$ for $cont$ whose kernel size is proportional to the distance between the current anchor point and the origin, $\parallel \Delta\bx^{anchor}_{m} \parallel$. That is $\exp{\big(-\dfrac{\parallel \Delta\bx_{ij}  - \Delta\bx^{anchor}_{m} \parallel^2}{2\sigma_{scaled}^2}\big)}$. Here $\sigma_{scaled} = \sigma + \beta\parallel \Delta\bx^{anchor}_{m} \parallel$ where $\sigma$ is the basic kernel size and $\beta$ is kernel rescale factor, a constant.
We developed this mechanism to help RFB to deal with relations at different scale, and we observe that it produces significantly better result than vanilla RBFs.
\end{itemize}

\subsection{Hyper-Parameter Selection  }\label{sec:parameters}
We perform grid search for all methods based on their performance on the validation sets.

\paragraph{Location Modeling} The hyper-parameters of   $theory$ models are  based on grid search  with $\fedim = (32, 64, 128, 256)$, $\pedim = (32, 64, 128, 256)$, $S = (4,8,16,32,64,128)$, and 
$\lambda_{min} = (1,5,10,50,100,200,500,1k)$ while $\lambda_{max} = 40k$ is decided based on the total size of the study area. 
We find out the best performances of different grid cell based models are obtained when $\fedim = 64$, $\pedim = 64$, $S = 64$, and 
$\lambda_{min} = 50$. 
%
In terms of $\model{tile}{loc}$, the hyper-parameters are selected from $c = (10, 50, 100, 200, 500, 1000)$ while $c=500$ gives us the best performance.
As for $\model{rbf}{loc}$, we do grid search on the hyper-parameters: $M = (10, 50, 100, 200, 400, 800)$ and $\sigma = (10^2, 10^3, 10^4, 10^5, 10^6, 10^7)$. The best performance of $\model{rbf}{loc}$ is obtain when $M = 200$ and $\sigma = 10^3$.
%
As for $\model{\aodha}{loc}$, grid search is performed on: $\numresnet = (1,2,3,4)$ and $\numneuron = (64,128,256,512)$ while $\numresnet = 3$ and $\numneuron = 512$ gives us the best result. 
%
All models use FFNs in their $\enc^{(x)}()$ except $\model{\aodha}{loc}$. The number of layers $f$ and the number of hidden state neurons $u$ of the FFN are selected from $f=(1,2,3)$ and $u=(128,256,512)$. We find out $f=1$ and $u=512$ give the best performance for $\model{\naive}{loc}$, $\model{tile}{loc}$, $\model{rbf}{loc}$, and $\model{theory}{loc}$. So we use them for every model for a fair comparison.

\paragraph{Spatial Context Modeling} 
Grid search is used for hyperparameter tuning and the best performance of different grid cell models is obtain when $\fedim = 64$, $\pedim = 64$, $S = 64$, and $\lambda_{min}=10$. We set $\lambda_{max} = 10k$ based on the maximum displacement between context points and center points to make the location encoding unique. 
As for multiple baseline models, grid search is used again to obtain the best model. The best model hyperparameters are shown in () besides the model names in Table \ref{tab:conteval}. Note that both $\model{rbf}{cont}$ and $\model{scaled\_rbf}{cont}$ achieve the best performance with $M=100$.



\subsection{Firing Pattern for the Neurons}\label{sec:firing}

\begin{figure*}[h!]
	\centering
	\begin{subfigure}[b]{0.49\textwidth}
		\centering
		\includegraphics[width=\textwidth]{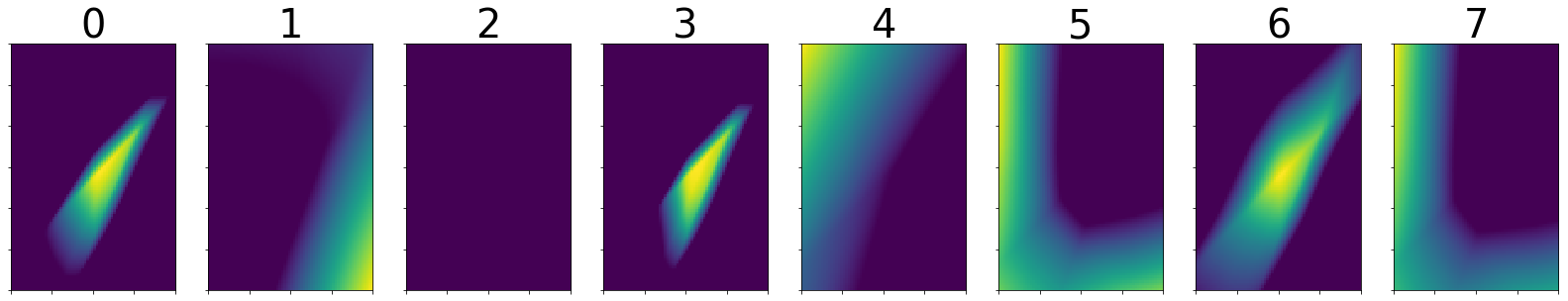}
		\subcaption[]{$\model{\naive}{loc}$}
		\label{fig:locnaive}
	\end{subfigure}
	\begin{subfigure}[b]{0.49\textwidth}
		\centering
		\includegraphics[width=\textwidth]{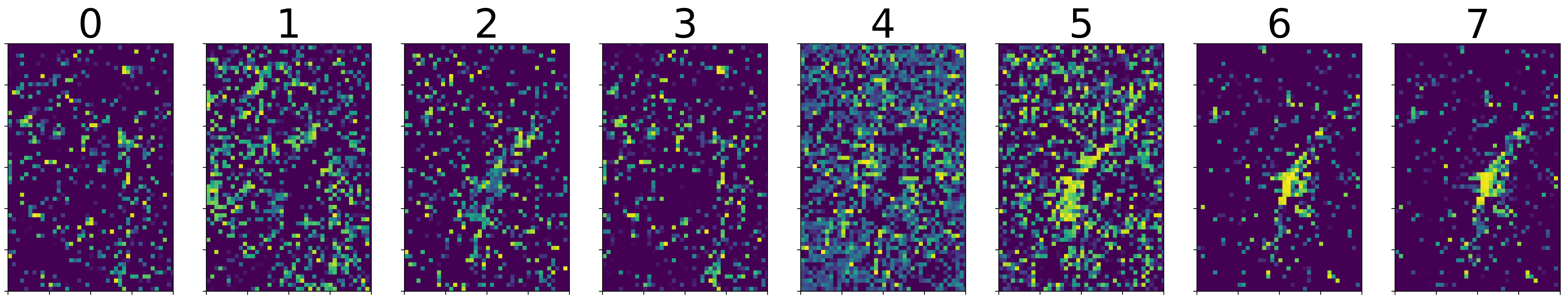}
		\subcaption[]{$\model{tile}{loc}$ ($c$=500)
		}
		\label{fig:loctile}
	\end{subfigure}
	\vspace*{-0.05cm}
	\begin{subfigure}[b]{0.49\textwidth}  
		\centering 
		\includegraphics[width=\textwidth]{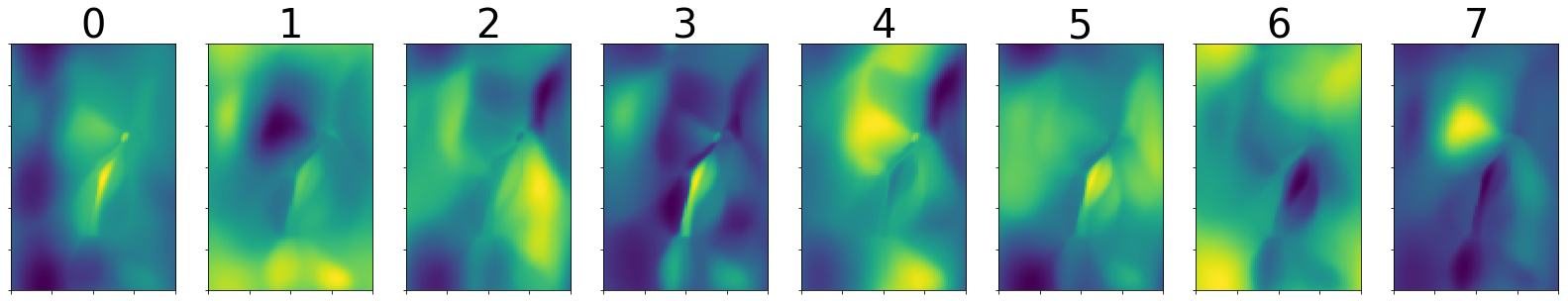}
		\subcaption[]{$\model{\aodha}{loc}$ 
		($\numresnet$=3,$\numneuron$=512)
		}
		\label{fig:locaodha}
	\end{subfigure}
	\vspace*{-0.05cm}
	\begin{subfigure}[b]{0.49\textwidth}
		\centering
		\includegraphics[width=\textwidth]{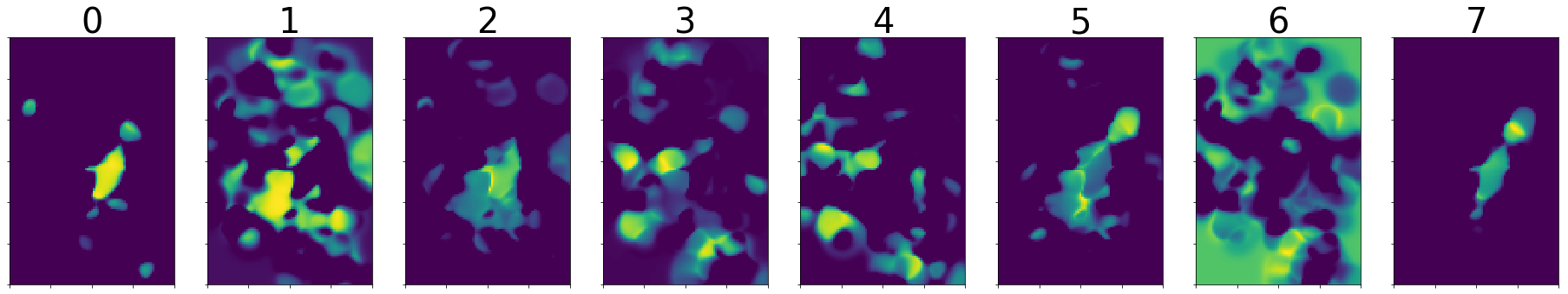}
		\subcaption[]{$\model{rbf}{loc}$ ($\sigma$=1k)
		}
		\label{fig:locrbf}
	\end{subfigure}
	\vspace*{-0.05cm}
	\begin{subfigure}[b]{0.49\textwidth}  
		\centering 
		\includegraphics[width=\textwidth]{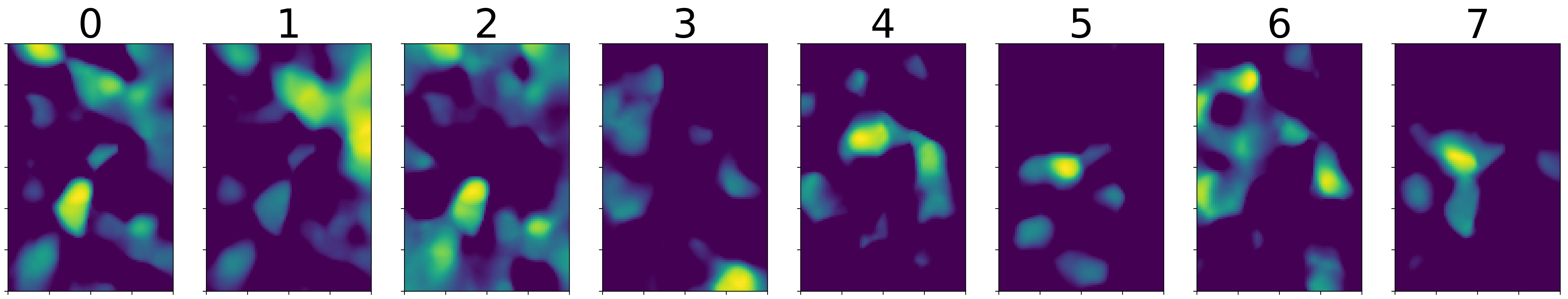}
		\subcaption[]{$\model{theory}{loc}$ ($\lambda_{min}$=1k)}
		\label{fig:loctheory1k}
	\end{subfigure}
	\vspace*{-0.05cm}
	\begin{subfigure}[b]{0.49\textwidth}  
		\centering 
		\includegraphics[width=\textwidth]{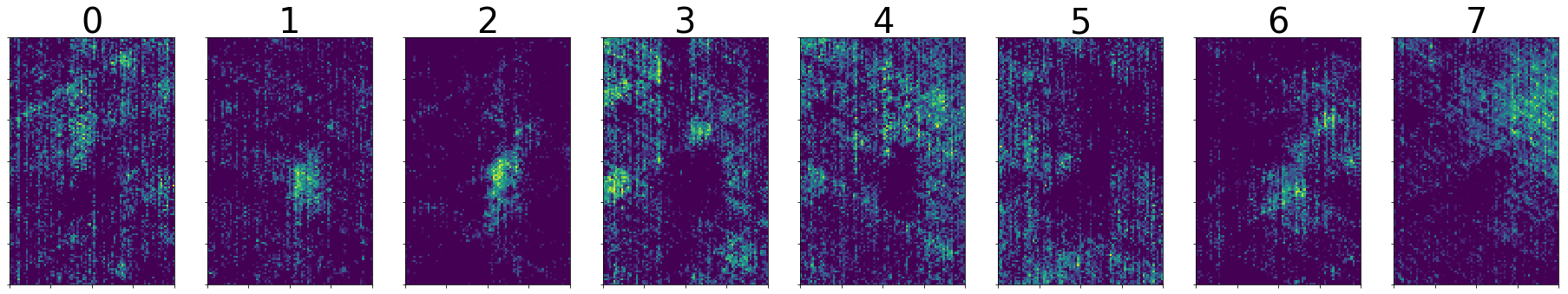}
		\subcaption[]{$\model{theory}{loc}$ ($\lambda_{min}$=50)}
		\label{fig:loctheory50}
	\end{subfigure}
	\vspace*{-0.05cm}
	\caption{\small
	The firing pattern for the first 8 neurons (out of 64) given different encoders in location modeling. 
	}
	\label{fig:locneuron}
	\vspace*{-0.45cm}
\end{figure*}

	
	

\newpage
\subsection{ Embedding clustering of RBF and theory models  }\label{sec:clustering}



\begin{figure*}[h!]
	\centering \tiny
	\vspace*{-0.2cm}
	\begin{subfigure}[b]{0.23\textwidth}  
		\centering 
		\includegraphics[width=\textwidth]{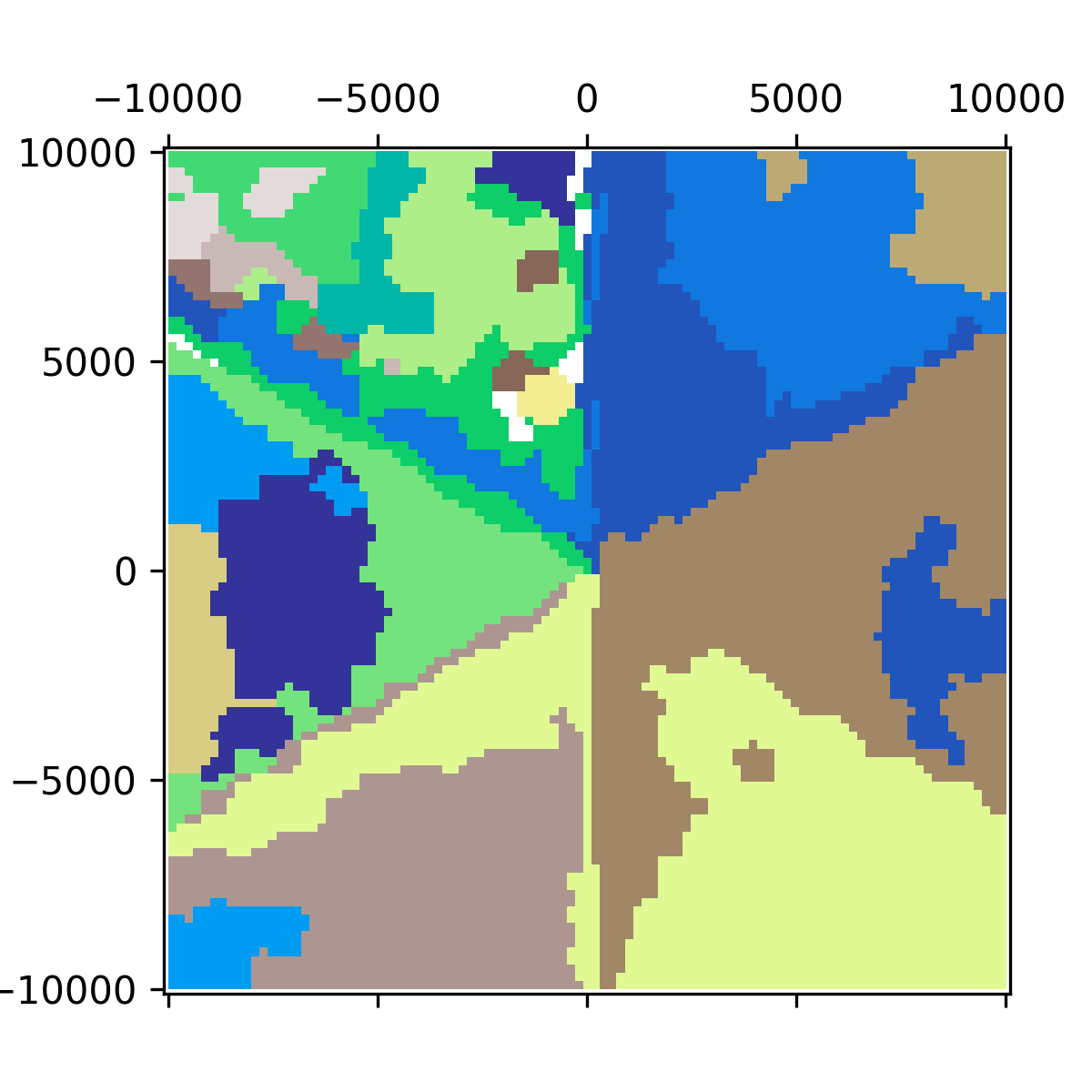}\vspace*{-0.2cm}
		\caption[]%
		{{\small  
		$\lambda_{min}$=200 
		}}    
		\label{fig:cont_theory_200}
	\end{subfigure}
	\hfill
	\begin{subfigure}[b]{0.23\textwidth}  
		\centering 
		\includegraphics[width=\textwidth]{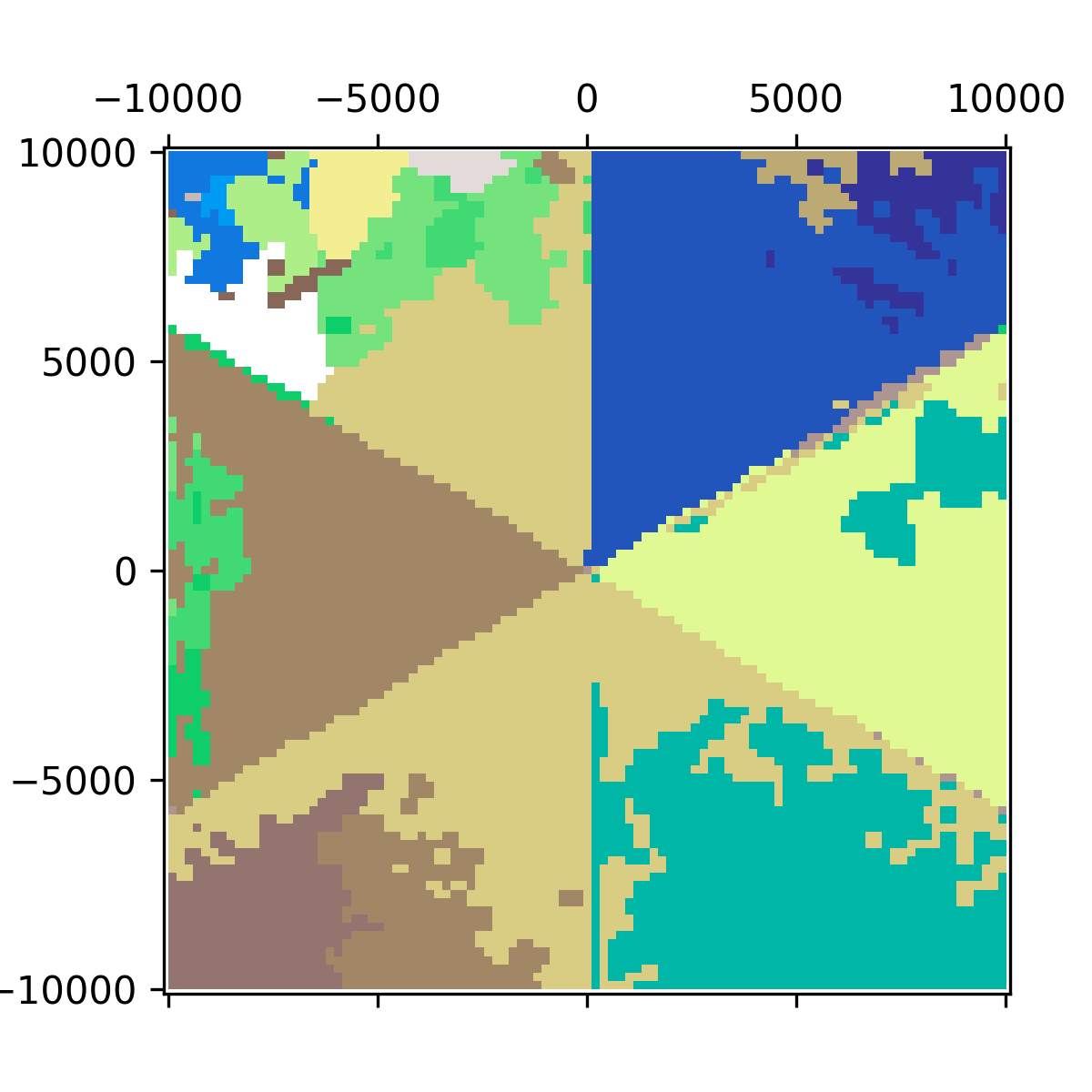}\vspace*{-0.2cm}
		\caption[]%
		{{\small  
		$\lambda_{min}$=100 
		}}    
		\label{fig:cont_theory_100}
	\end{subfigure}
	\hfill
	\begin{subfigure}[b]{0.23\textwidth}  
		\centering 
		\includegraphics[width=\textwidth]{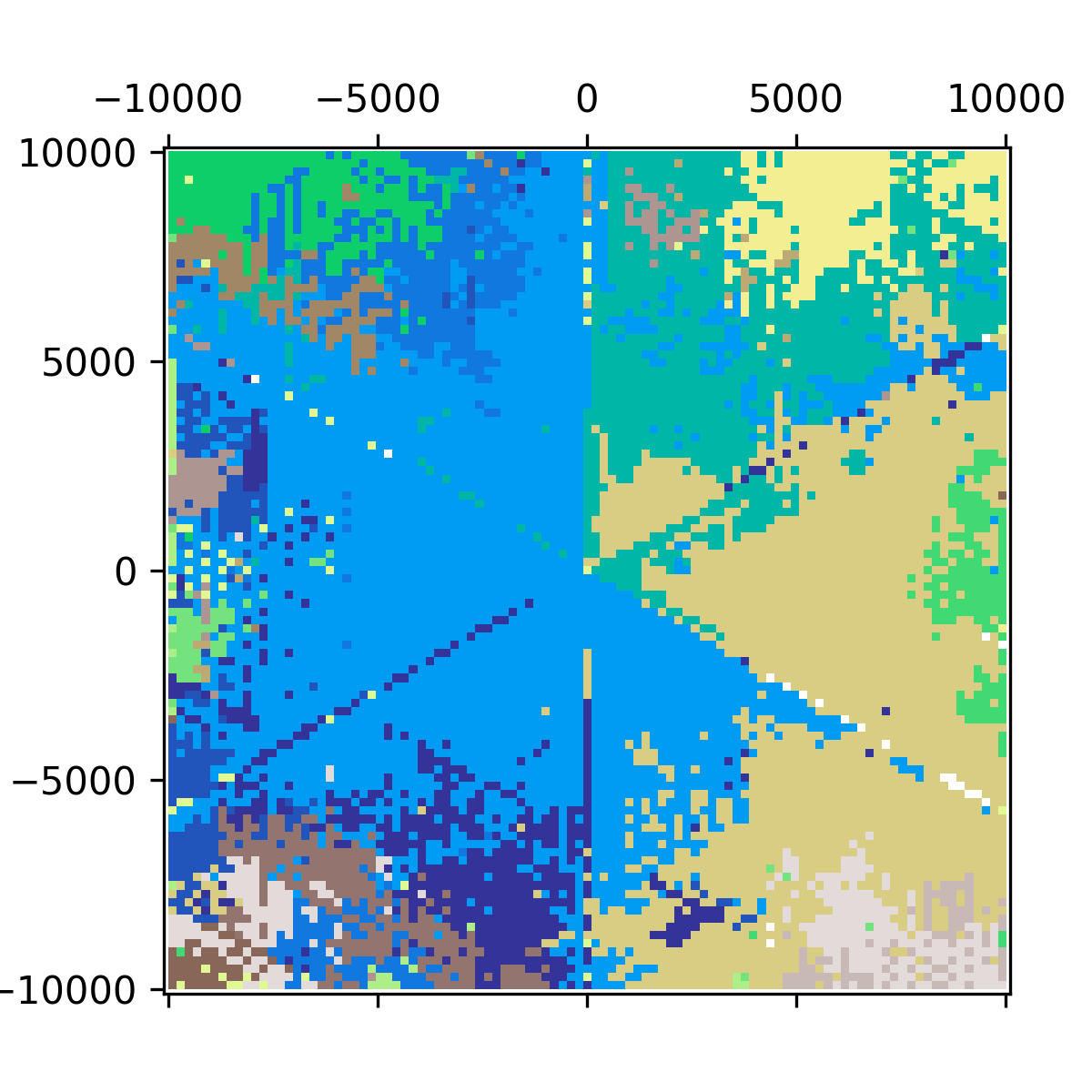}\vspace*{-0.2cm}
		\caption[]%
		{{\small  
		$\lambda_{min}$=50 
		}}    
		\label{fig:cont_theory_50}
	\end{subfigure}
	\hfill
	\begin{subfigure}[b]{0.23\textwidth}  
		\centering 
		\includegraphics[width=\textwidth]{./fig/cont_theory_64_10k_10_1h512e64.png}\vspace*{-0.2cm}
		\caption[]%
		{{\small 
		$\lambda_{min}$=10
		}}    
		\label{fig:cont_theory_10}
	\end{subfigure}
	\hfill
	\begin{subfigure}[b]{0.23\textwidth}  
		\centering 
		\includegraphics[width=\textwidth]{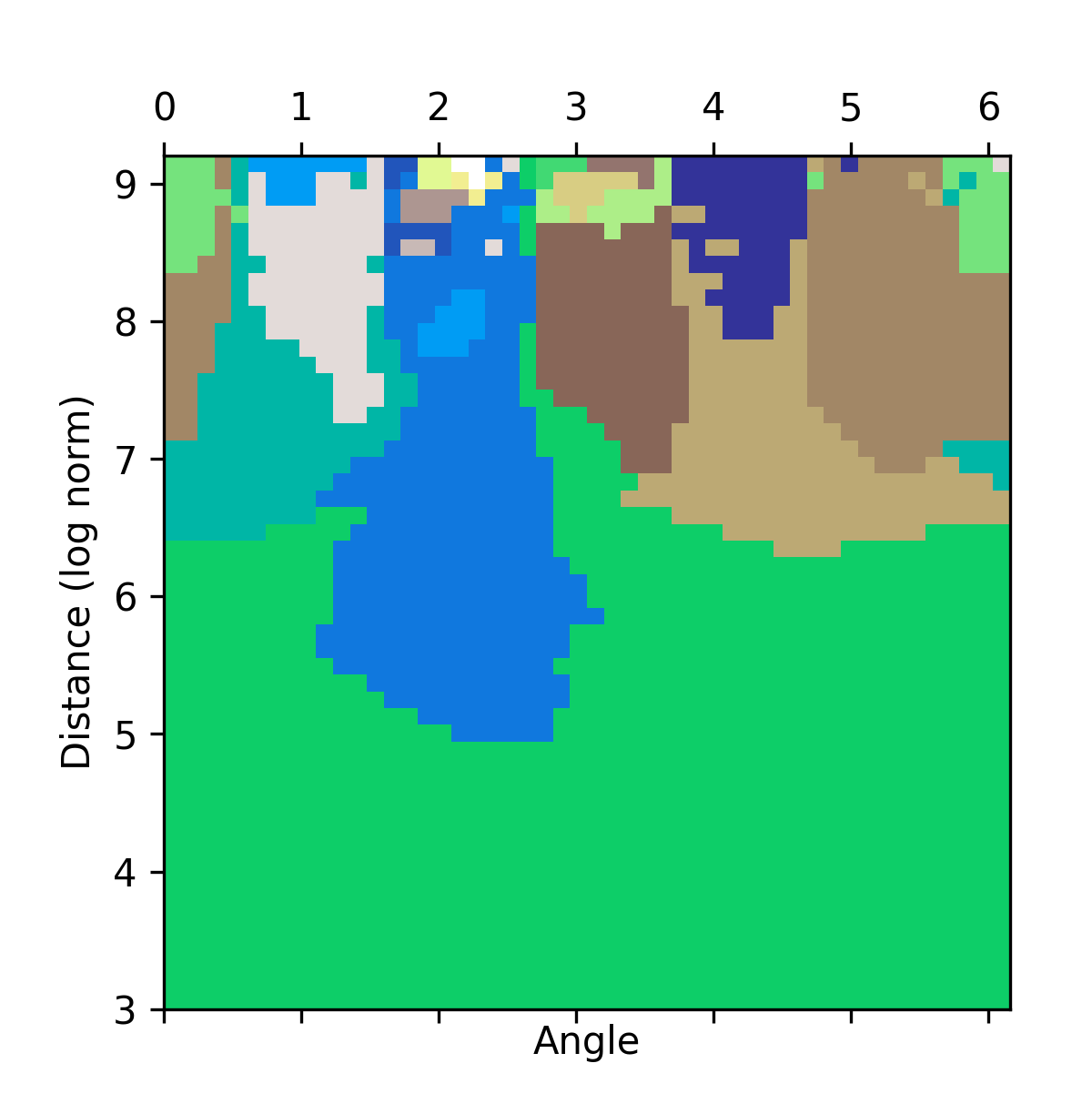}\vspace*{-0.2cm}
		\caption[]%
		{{\small  
		$\lambda_{min}$=200 
		}}    
		\label{fig:cont_theory_200_polar}
	\end{subfigure}
	\hfill
	\begin{subfigure}[b]{0.23\textwidth}  
		\centering 
		\includegraphics[width=\textwidth]{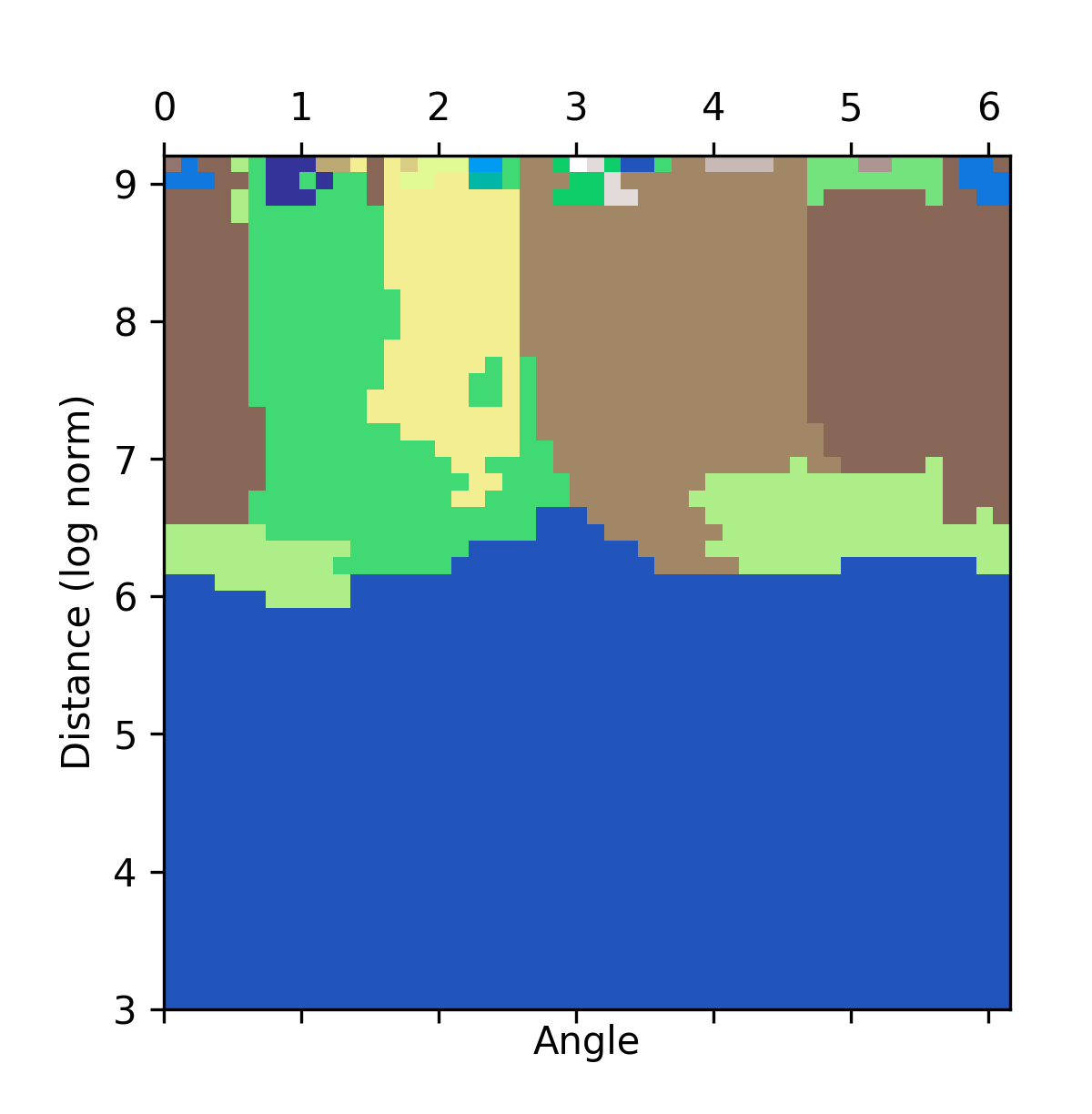}\vspace*{-0.2cm}
		\caption[]%
		{{\small  
		$\lambda_{min}$=100 
		}}    
		\label{fig:cont_theory_100_polar}
	\end{subfigure}
	\hfill
	\begin{subfigure}[b]{0.23\textwidth}  
		\centering 
		\includegraphics[width=\textwidth]{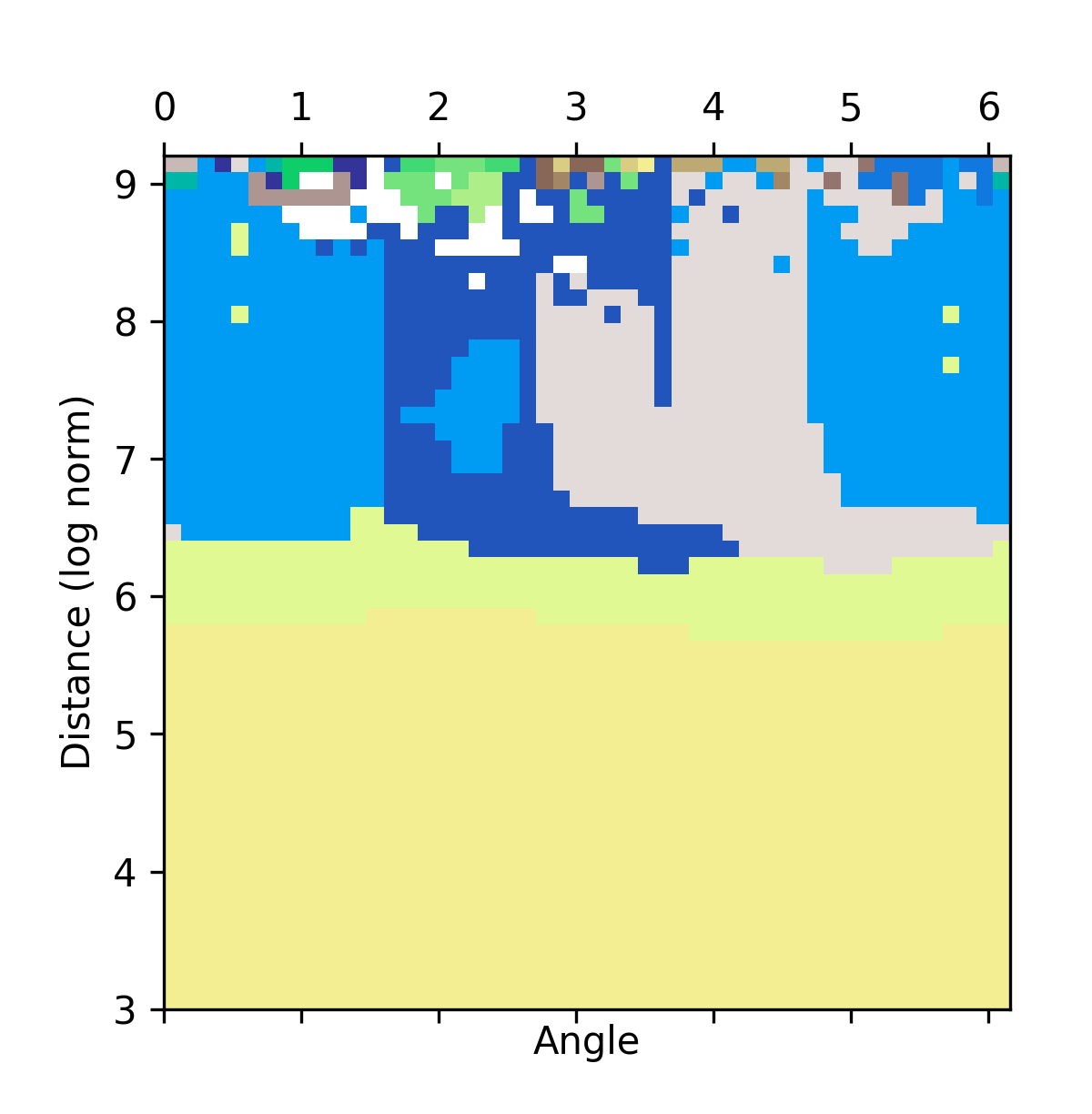}\vspace*{-0.2cm}
		\caption[]%
		{{\small  
		$\lambda_{min}$=50 
		}}    
		\label{fig:cont_theory_50_polar}
	\end{subfigure}
	\hfill
	\begin{subfigure}[b]{0.23\textwidth}  
		\centering 
		\includegraphics[width=\textwidth]{./fig/cont_theory_64_10k_10_1h512e64-polar.png}\vspace*{-0.2cm}
		\caption[]%
		{{\small 
		$\lambda_{min}$=10
		}}    
		\label{fig:cont_theory_10_polar}
	\end{subfigure}
	\caption{
	\small
	Embedding clustering in the original space of (a)(b)(c)(d) $\model{theory}{cont}$ with different $\lambda_{min}$, but the same $\lambda_{max}=10k$ and $\nscale = 64$. (e)(f)(g)(h) are the embedding clustering results of the same models in the polar-distance space. All models use 1 hidden ReLU layers of 512 neurons. 
	} 
	\label{fig:poicl}
\end{figure*}

\begin{figure*}[h!]
	\centering \tiny
	\vspace*{-0.2cm}
	\begin{subfigure}[b]{0.23\textwidth}  
		\centering 
		\includegraphics[width=\textwidth]{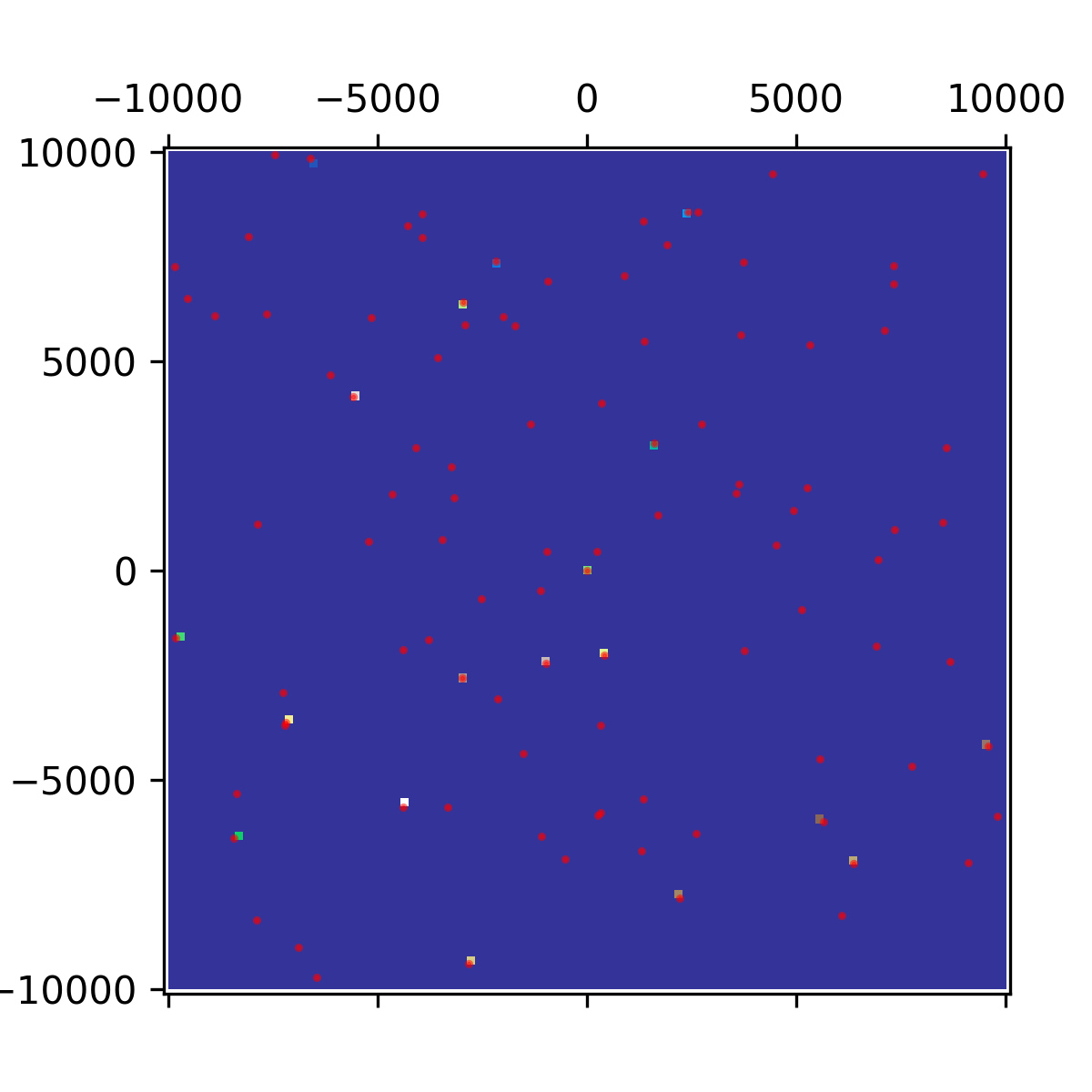}\vspace*{-0.2cm}
		\caption[]%
		{{\small 
		$\beta$=0.0
		}}    
		\label{fig:cont_rbf_00}
	\end{subfigure}
	\hfill
	\begin{subfigure}[b]{0.235\textwidth}  
		\centering 
		\includegraphics[width=\textwidth]{./fig/{cont_rbf_n100_10_0.1_1h512e64}.png}\vspace*{-0.03cm}
		\caption[]%
		{{\small  
		$\beta$=0.1 
		}}    
		\label{fig:cont_rbf_01}
	\end{subfigure}
	\hfill
	\begin{subfigure}[b]{0.23\textwidth}  
		\centering 
		\includegraphics[width=\textwidth]{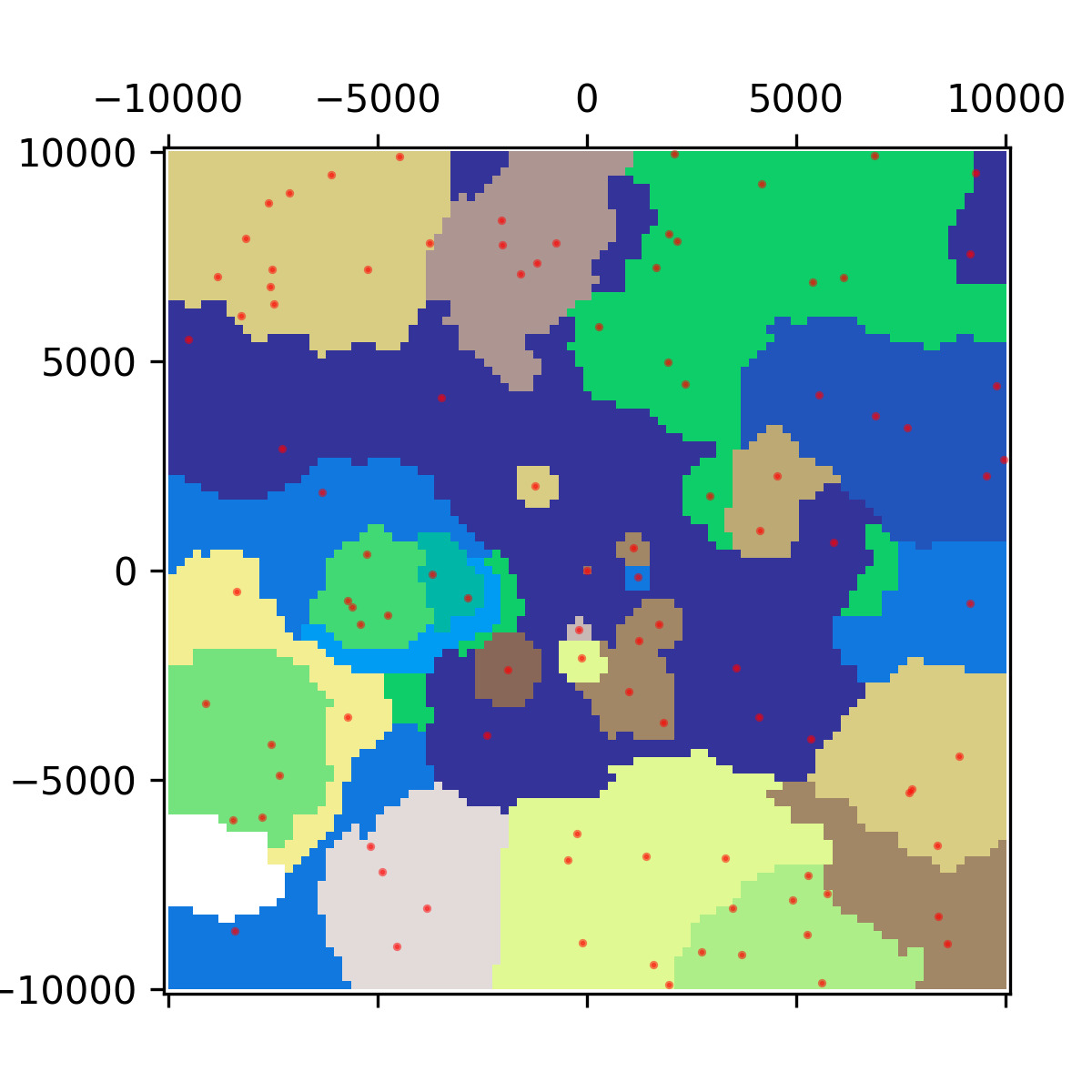}\vspace*{-0.2cm}
		\caption[]%
		{{\small  
		$\beta$=0.2 
		}}    
		\label{fig:cont_rbf_02}
	\end{subfigure}
	\hfill
	\begin{subfigure}[b]{0.23\textwidth}  
		\centering 
		\includegraphics[width=\textwidth]{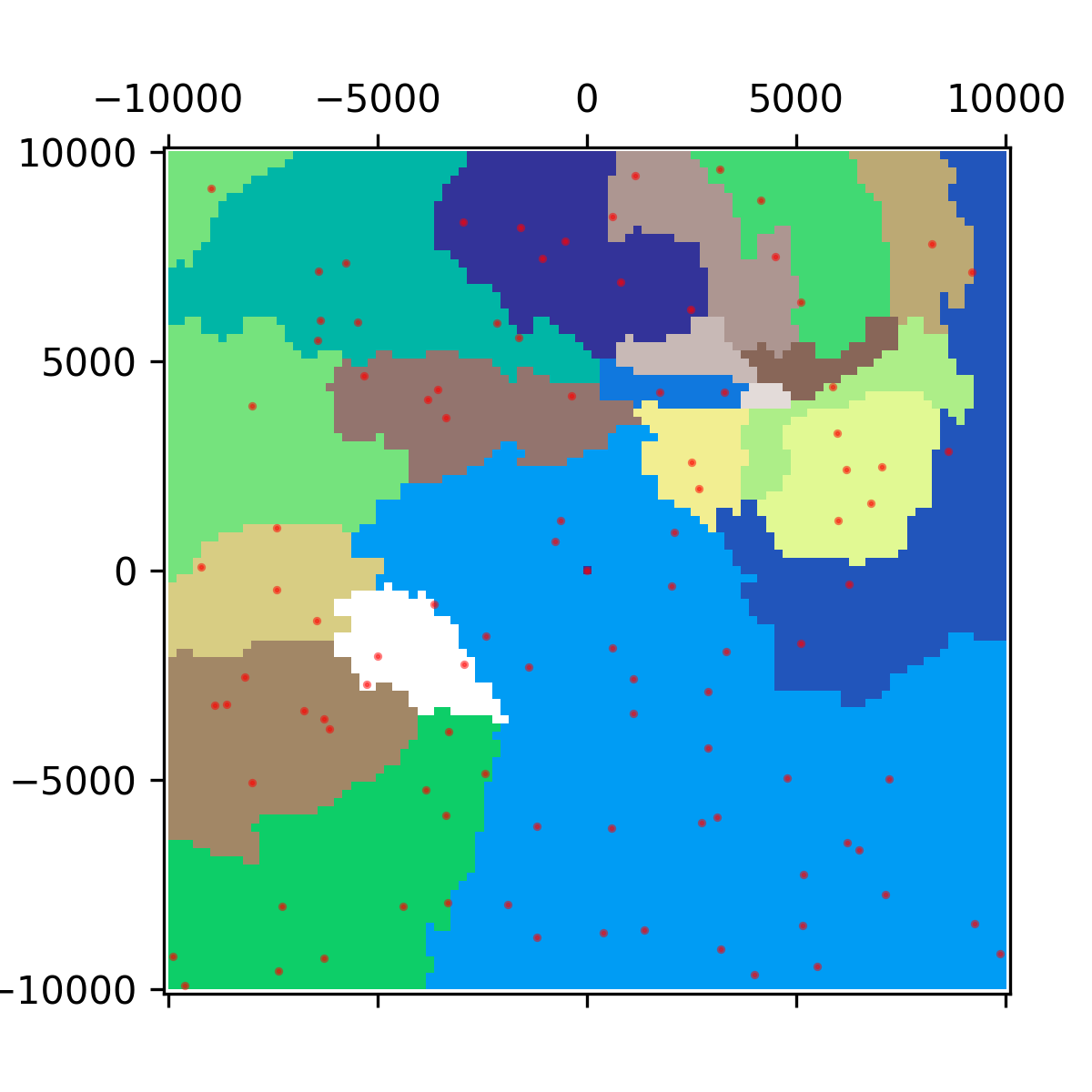}\vspace*{-0.2cm}
		\caption[]%
		{{\small 
		$\beta$=0.3
		}}    
		\label{fig:cont_rbf_03}
	\end{subfigure}
	\begin{subfigure}[b]{0.23\textwidth}  
		\centering 
		\includegraphics[width=\textwidth]{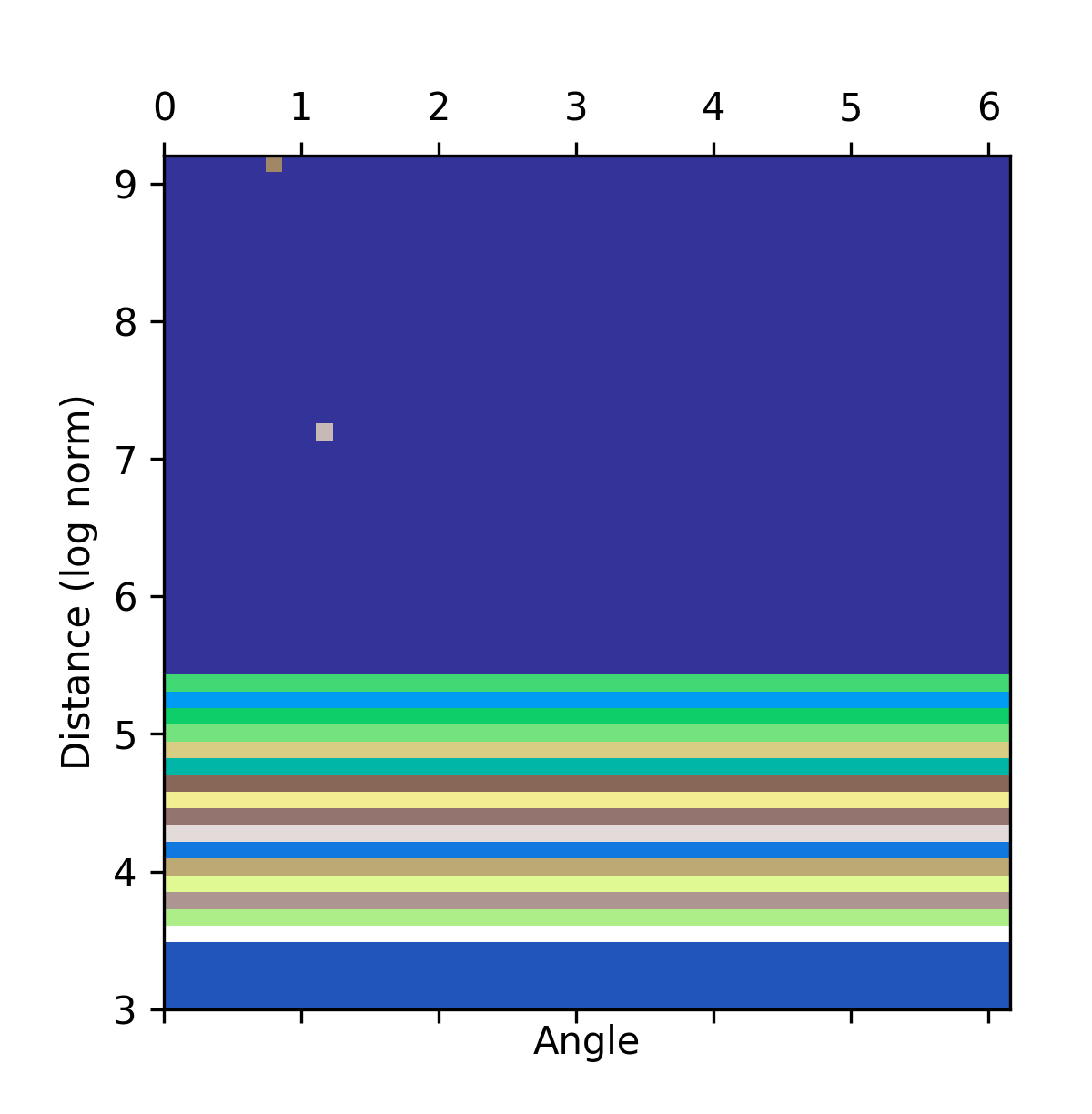}\vspace*{-0.2cm}
		\caption[]%
		{{\small 
		$\beta$=0.0
		}}    
		\label{fig:cont_rbf_00_polar}
	\end{subfigure}
	\hfill
	\begin{subfigure}[b]{0.22\textwidth}  
		\centering 
		\includegraphics[width=\textwidth]{./fig/{cont_rbf_n100_10_0.1_1h512e64-polar}.png}\vspace*{-0.2cm}
		\caption[]%
		{{\small  
		$\beta$=0.1 
		}}    
		\label{fig:cont_rbf_01_polar}
	\end{subfigure}
	\hfill
	\begin{subfigure}[b]{0.23\textwidth}  
		\centering 
		\includegraphics[width=\textwidth]{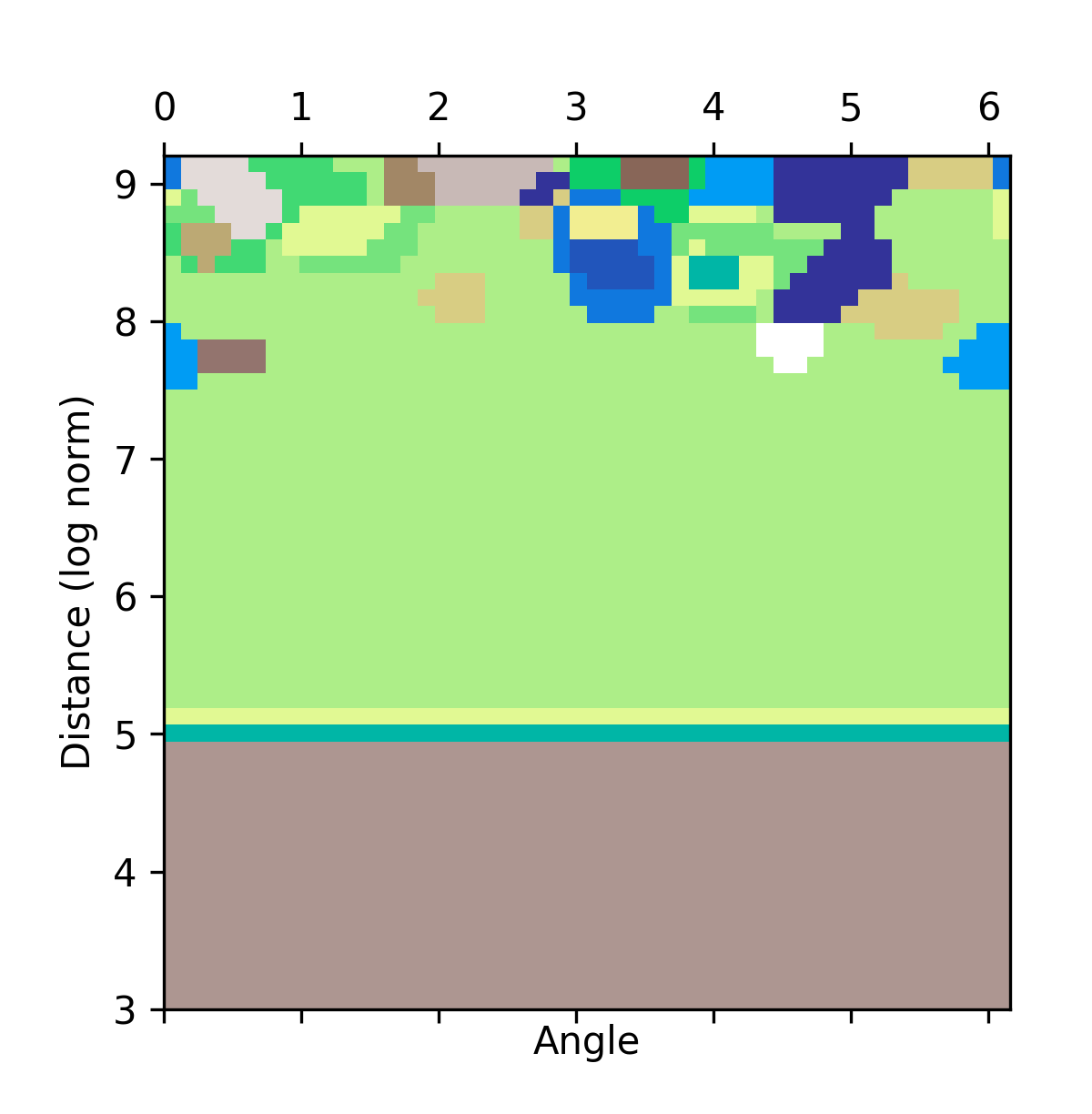}\vspace*{-0.2cm}
		\caption[]%
		{{\small  
		$\beta$=0.2 
		}}    
		\label{fig:cont_rbf_02_polar}
	\end{subfigure}
	\hfill
	\begin{subfigure}[b]{0.23\textwidth}  
		\centering 
		\includegraphics[width=\textwidth]{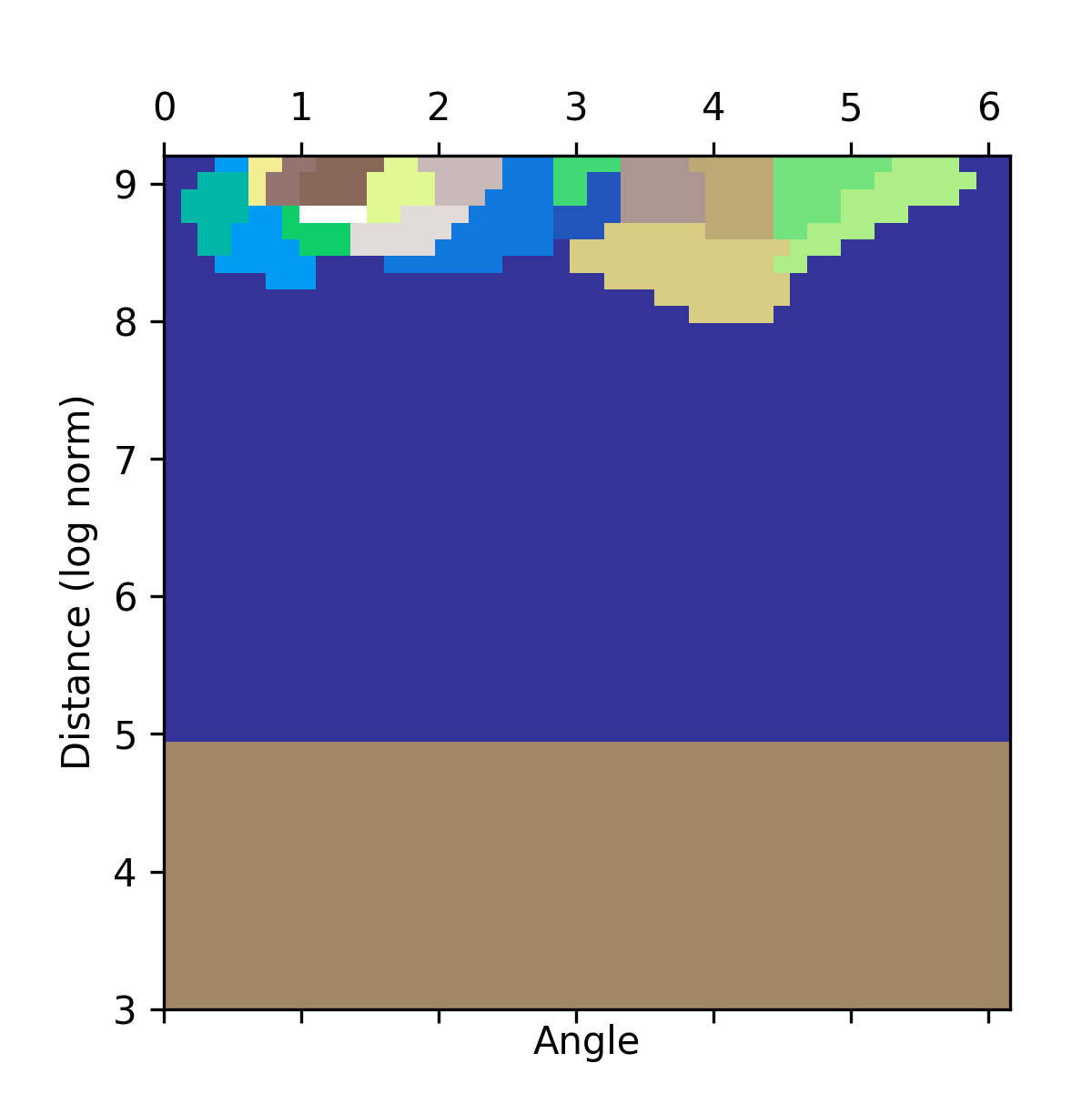}\vspace*{-0.2cm}
		\caption[]%
		{{\small 
		$\beta$=0.3
		}}    
		\label{fig:cont_rbf_03_polar}
	\end{subfigure}
	\caption{
	\small
	Embedding clustering of RBF models with different kernel rescalar factor $\beta$ (a)(b)(c)(d) in the original space; (e)(f)(g)(h) in the polar-distance space. Here $\beta$=0.0 indicates the original RBF model. All models use $\sigma$=10m as the basic kernel size and 1 hidden ReLU layers of 512 neurons. 
	} 
	\label{fig:poicl}
\end{figure*}

\end{document}